\documentclass[letterpaper, 10 pt, journal, twoside]{IEEEtran}  %

\IEEEoverridecommandlockouts                              %

\usepackage{times}
\usepackage{textcomp}

\usepackage[numbers, sort&compress]{natbib}
\usepackage{multicol}
\usepackage[bookmarks=true]{hyperref}
\usepackage{xurl}
\usepackage{xspace}
\usepackage{amsmath}
\usepackage{graphicx}
\usepackage{enumitem}
\usepackage[capitalise, nameinlink]{cleveref}
\usepackage{booktabs}
\usepackage[font=small, labelfont=bf]{caption}
\usepackage{makecell}
\usepackage{multirow}
\usepackage{bm}
\usepackage{amssymb}
\usepackage{tcolorbox}
\usepackage{subcaption}
\usepackage{algorithm}
\usepackage{algpseudocode}

\hypersetup{
    colorlinks=true,
    linkcolor=orange,
    filecolor=magenta,      
    urlcolor=orange,
    citecolor=orange,
    pageanchor=false,
}

\newcommand{\method}{RADAR\xspace}

\begin{document}

\title{Grounding Robot Generalization in Training Data via\\[-4pt] Retrieval-Augmented VLMs}

\author{Jensen Gao$^{1,2}$,
Dorsa Sadigh$^{1}$,
Sandy Huang$^{2}$,
Dhruv Shah$^{2,3}$
\vspace{-24pt}
}%

\maketitle

\markboth{IEEE Robotics and Automation Letters. Preprint Version. Accepted August, 2026}
{Gao \MakeLowercase{\textit{et al.}}: Grounding Robot Generalization in Training Data via Retrieval-Augmented VLMs}

\begin{abstract}
Recent work on robot manipulation has advanced policy generalization to novel scenarios. However, it is often difficult to characterize how different evaluation settings actually represent generalization from the training distribution of a given policy. To work towards more precise evaluation of generalization in robotics, we propose \method, a scalable framework for directly comparing test-time evaluation tasks to policy training data, to determine what form of policy generalization is required. \method consists of a two-stage pipeline: first, retrieval using generalist policy embeddings identifies which training examples are relevant for a given evaluation task. Next, vision-language models (VLMs) analyze the evaluation task against the retrieved data, outputting interpretable analysis on how they compare along a variety of axes, and an overall classification of what type of policy generalization is required. Through controlled experiments, we demonstrate that VLMs are effective at analyzing data for generalization, and that our retrieval step effectively identifies examples needed to make accurate classifications with respect to the training data. Furthermore, we scale \method to large-scale datasets, where we observe agreement with human-defined benchmark conditions from prior work. We provide demonstrations at our website \href{https://radar-analysis.github.io}{radar-analysis.github.io}.
\end{abstract}

\begin{IEEEkeywords}
Big Data in Robotics and Automation, Software Tools for Benchmarking and Reproducibility, Deep Learning in Grasping and Manipulation.
\end{IEEEkeywords}

\renewcommand{\thefootnote}{}
\footnotetext{
Manuscript received: February 22, 2026; Revised: June 23, 2026; Accepted: August 10, 2026.

This paper was recommended for publication by Editor Wei Pan upon evaluation of the Associate Editor and Reviewers' comments.

Jensen Gao is with $^{1}$Stanford University, and did this work as a student researcher at $^{2}$Google DeepMind. Dorsa Sadigh is with $^{1}$Stanford University. Sandy Huang is with $^{2}$Google DeepMind. Dhruv Shah is with $^{2}$Google DeepMind and $^{3}$Princeton University. Correspondence to: {\tt\footnotesize jenseng@stanford.edu}.

Digital Object Identifier (DOI): see top of this page.
}
\renewcommand{\thefootnote}{\arabic{footnote}}

\section{Introduction}
\label{sec:introduction}
\IEEEPARstart{A}{chieving} broad generalization is one of the most important and challenging problems in robotics. We want robots that can be deployed in unseen, open-world scenarios, such as a robot that can do the dishes in brand new kitchens. But measuring progress is difficult: the range of settings we care about is immense, yet we can only evaluate a narrow set of conditions in practice. It thus becomes important to characterize how evaluations capture generalization from training data, to better understand policy capabilities and limitations.

Recent work has claimed that generalist robot policies can generalize beyond their training data. %
However, real-world benchmarking in robotics is notoriously difficult to reproduce, and models are increasingly trained on large-scale proprietary datasets~\cite{team2025gemini, team2025gemini1.5, black2024pi_0, intelligence2025pi}. As a result, it is often unclear to what extent evaluations actually capture generalization from training data. %
When a robot successfully does the dishes in a new kitchen, how much generalization is actually happening? Simply designating the task as ``unseen'' is vague, as the extent of generalization can vary greatly: a significantly different kitchen layout involves a greater degree of generalization than only a change in countertop color.

To work towards better understanding policy generalization, we propose \textbf{R}etrieval-\textbf{A}ugmented \textbf{D}ata \textbf{A}nalysis for \textbf{R}obotics (\textbf{\method}), a simple and scalable framework for characterizing how evaluation conditions constitute generalization from robot training data. \method uses vision-language models (VLMs) to directly compare training data to evaluation tasks, by prompting the VLM with task observations and language instructions as well as a taxonomy to structure its analysis along different generalization axes~\cite{taxonomy2025arxiv}. Then, it outputs qualitative, human-interpretable descriptions of how the evaluation task deviates from the training data along these axes, as well as an overall classification of the evaluation task as either \textbf{in-distribution (ID)}, or requiring \textbf{visual} or \textbf{behavioral} generalization.

To scale to large robot datasets, we first use embedding-based retrieval to identify relevant examples from training data, and use the VLM to only analyze these. For our embeddings, we propose using internal representations from generalist vision-language-action (VLA) policies trained on large-scale robot data. These embeddings capture visual and semantic invariances from their internet-scale pretraining, while being sensitive to scene changes that require new behavior from their exposure to large-scale robot data. These are key properties for effective retrieval in our framework. Importantly, these embeddings can be applied to data unseen by the policy.

We evaluate \method through controlled experiments with diverse data mixtures across three task families on the ALOHA 2 platform~\cite{aldaco2024aloha}. In these experiments, we show that generalist policy embeddings perform well at identifying training examples needed to make accurate downstream classifications, achieving recall rates of 80--90\% on such examples when retrieving only 5--10\% of the original dataset. We also show that VLMs can be effective at comparing task instances, achieving 92.5\% accuracy on classifying overall policy generalization for one task family. However, we observe that modern VLMs can still be limited when analyzing tasks with more subtle visual differences, resulting in worse performance. %

Lastly, we scale \method to two large-scale datasets: Bridge V2~\cite{walke2023bridgedata}, and a dataset of over 1M ALOHA 2 demonstrations. We use it to assess evaluation tasks used in prior work for benchmarking generalist manipulation. We show that \method often produces analysis in agreement with human-defined characterizations of generalization, while also revealing cases where these characterizations may have subtle limitations.

\section{Related Work}
\label{sec:related_work}
\begin{figure*}[t]
    \centering
    \includegraphics[width=\linewidth]{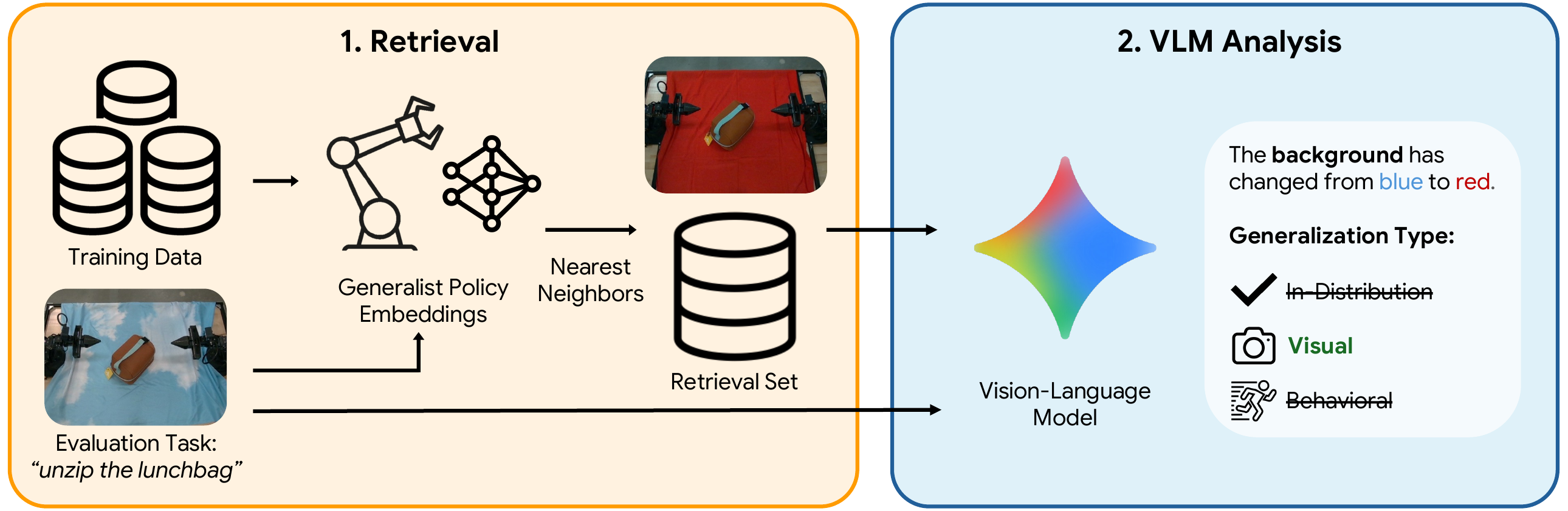}
    \caption{\small We outline the two-stage pipeline of \method. Given an evaluation task, we first identify relevant examples from training data using nearest neighbors with embeddings from a generalist robot policy. Next, we use vision-language models to analyze the evaluation task against the retrieved examples to categorize what generalization it represents w.r.t.\ the training data: \textbf{in-distribution (ID)}, \textbf{visual}, or \textbf{behavioral}.}
    \label{fig:method_overview}
    \vspace{-10pt}
\end{figure*}

We review related work on generalization in robotics, vision-language models, and retrieval in robotics.

\smallskip \noindent \textbf{Generalization in Robotics.}
Many efforts have worked towards generalization in robotics via scaling data~\cite{dasari2019robonet, walke2023bridgedata, fang2023rh20t, o2023open, khazatsky2024droid} and models~\cite{kim24openvla, team2025gemini, team2025gemini1.5, intelligence2025pi}, and investigating best practices for generating robot data~\cite{gao2024efficient, lin2024data, saxena2025what}. When evaluating generalist policies, simulation can make evaluations more standardized and efficient~\cite{liu2023libero, li2024evaluating, jain2025polaris}, but faces challenges including limited diversity and the sim-to-real gap. To evaluate for real-world generalization, different works largely use their own protocols, making results difficult to interpret and reproduce. Some works aim to mitigate this, e.g., by proposing a taxonomy of generalization axes~\cite{taxonomy2025arxiv} or a crowd-sourced benchmarking platform~\cite{atreya2025roboarena}. However, how an evaluation represents generalization is typically decided by ad hoc human judgment specific to each benchmark, which limits rigor and reproducibility. In our work, we propose a data analysis framework to systematically characterize generalization required by evaluations.%

\smallskip \noindent \textbf{Vision-Language Models.}
VLMs have emerged as powerful, general-purpose tools for visual reasoning. VLMs have previously been used for embodied reasoning, including for robot planning~\cite{driess2023palm}, physical scene understanding~\cite{gao2024physically}, spatial reasoning~\cite{chen2024spatialvlm}, and reasoning over execution failures~\cite{duan2025aha}. In our work, we leverage VLMs to compare scenes in robot data for characterizing generalization conditions.

\smallskip \noindent \textbf{Retrieval and Embeddings.}
Data retrieval for robotics has been studied for identifying training data that most benefits a downstream task, using embedding-based similarity~\cite{nasiriany2022learning, du2023behavior, lin2024flowretrieval, memmel2025strap, xie2025data}. In our setting, we focus on identifying relevant portions of training data to analyze, not for training. Also, unlike these works, we use embeddings from robot policies, which are especially suited for our setting. Policy embeddings have been used to flag out-of-distribution (OOD) inputs and anticipate failures~\cite{majumdar2025predictive, zha2025guidingdatacollectionfactored, gu2025safe}, giving a binary, policy-specific verdict at deployment. We instead retrieve examples for VLM analysis (a form of RAG~\cite{lewis2020retrieval}), which yields a dataset-relative characterization of generalization, independent of policy success.

\section{Problem Formulation}
\label{sec:setup}
We develop our problem formulation based on $\bigstar$-Gen, a taxonomy for characterizing generalization based on task perturbations~\cite{taxonomy2025arxiv}. In particular, we consider how changes to tasks affect the input and output modalities of a visual-lingual control policy $\pi(a \mid o, l)$, which takes in an image observation $o$, a language instruction $l$, and outputs a distribution over actions $a$. We consider \textbf{visual} task perturbations that change the initial distribution of observations $o$, and \textbf{behavioral} perturbations that change the optimal distribution over actions $a$ (see Section III in~\cite{taxonomy2025arxiv} for more details). Unlike in~\cite{taxonomy2025arxiv}, we do not focus on \textbf{semantic-only} perturbations to instructions, to simplify our analysis and focus on what we consider more challenging aspects of robot generalization (e.g., language models alone can more easily address \textbf{semantic-only} generalization).

\begin{figure*}[t]
    \centering
    \includegraphics[width=\linewidth]{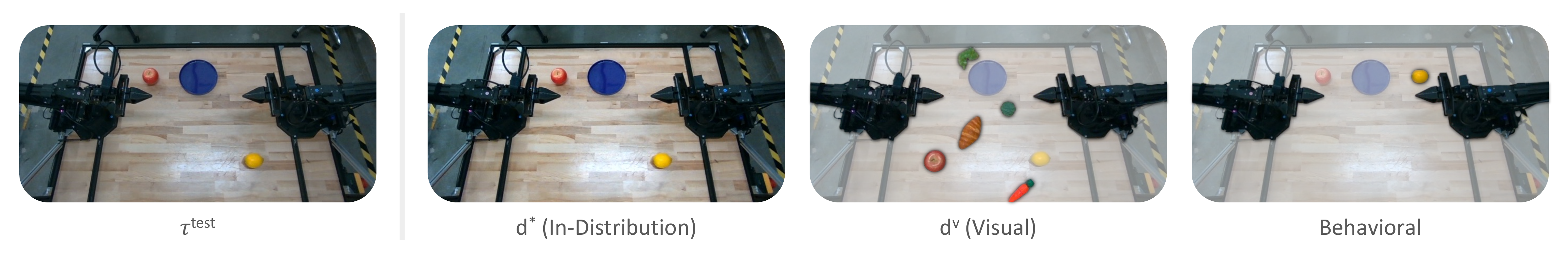}
    \caption{\small (Left) Test example $\tau^{\text{test}}$ for the task \emph{``put the lemon on the plate''}. (Right) If $d^*$ (e.g., minor lighting change) is found in $\mathcal{D}$, then $\tau^{\text{test}}$ is \textbf{in-distribution (ID)}. Otherwise, if $d^v$ (e.g., distractor objects) is found in $\mathcal{D}$, then this is \textbf{visual} generalization. If neither case applies, then all $d \in \mathcal{D}$ involve different optimal behavior than $\tau^{\text{test}}$ (e.g., changed object poses), and this is \textbf{behavioral} generalization.}
    \label{fig:gen_examples}
    \vspace{-10pt}
\end{figure*}

We assume a dataset $\mathcal{D} = (d^1, \dots, d^N)$ of $N$ demonstrations. Each $d^i = \{(o^i_0, a^i_0), \dots, (o^i_{T^i - 1}, a^i_{T^i - 1}), l^i\}$ is a sequence of $T^i$ observation-action pairs and a language instruction $l^i$. We then have a test-time evaluation task $\tau^{\text{test}}$, represented by an initial observation distribution $p_{\text{test}}(o_0)$ and language instruction $l^{\text{test}}$. In practice, we assume a sample $o^{\text{test}}_0 \sim p_{\text{test}}(o_0)$ rather than the full distribution, and that $p_{\text{test}}(o_0)$ is narrow (e.g., minor object pose variation, lighting).

We characterize how $\tau^{\text{test}}$ relates to a demonstration $d \in \mathcal{D}$ in two ways: (i) whether it is a \emph{visual match}, i.e., the initial observation in $d$ is ID w.r.t.\ $p_{\text{test}}(o_0)$, and (ii) whether it is a \emph{behavioral match}, i.e., the action sequence in $d$ is optimal behavior for $\tau^{\text{test}}$. These correspond to the absence of \textbf{visual} and \textbf{behavioral} perturbations, respectively. The relationship of $\tau^{\text{test}}$ to $\mathcal{D}$ is then determined as follows:
\begin{itemize}
    \item $\tau^{\text{test}}$ is \textbf{in-distribution (ID)} w.r.t.\ $\mathcal{D}$ if there exists $d^* \in \mathcal{D}$ that is both a \emph{visual} and \emph{behavioral match} to it.
    \item $\tau^{\text{test}}$ captures \textbf{visual} generalization w.r.t.\ $\mathcal{D}$ if no such $d^*$ exists, but there exists $d^v \in \mathcal{D}$ that is a \emph{behavioral match} (i.e., the initial observation $o^v_0$ is OOD w.r.t.\ $p_{\text{test}}(o_0)$, but the action sequence in $d^v$ is optimal behavior for $\tau^{\text{test}}$).
    \item $\tau^{\text{test}}$ captures \textbf{behavioral} generalization w.r.t.\ $\mathcal{D}$ if no $d \in \mathcal{D}$ is a \emph{behavioral match}. This could be due to changes in the scene or language specification (note that $d$ could be a \emph{visual match}, but not a \emph{behavioral match} due to different language instructions).
\end{itemize}

\smallskip \noindent In summary, our objective is to classify $\tau^{\text{test}}$ as either \textbf{ID} to $\mathcal{D}$, or capturing \textbf{visual} or \textbf{behavioral} generalization. We visualize an example of each scenario in \cref{fig:gen_examples}.

We note that $\mathcal{D}$ can be any dataset one desires to assess generalization w.r.t.\ (e.g., the entirety of a policy's pre-training data, or a more narrow post-training dataset).

\section{Retrieval-Augmented Data Analysis}
\label{sec:method}
In this section, we outline our proposed framework \method, which consists of two stages (retrieval and VLM analysis). We visualize an overview of \method in \cref{fig:method_overview}.

\subsection{Retrieval via Embeddings}
In order for \method to be scalable to large datasets $\mathcal{D}$, we first use retrieval to identify a sub-dataset $\mathcal{D}_{\text{retrieval}} \subset \mathcal{D}$, where $|\mathcal{D}_{\text{retrieval}}| \ll N$, to reduce the amount of data to be analyzed by the VLM. Our goal is to choose $\mathcal{D}_{\text{retrieval}}$ such that the VLM can produce an accurate classification of $\tau^{\text{test}}$ w.r.t.\ the original dataset $\mathcal{D}$, i.e., $\tau^{\text{test}}$ has the same generalization relationship to $\mathcal{D}_{\text{retrieval}}$ as to $\mathcal{D}$.

To obtain $\mathcal{D}_{\text{retrieval}}$, we select the $k$ nearest neighbors of $\tau^{\text{test}}$ in some embedding space induced by an embedding function $\phi(o_0, l)$, which takes in initial observations and language instructions. Formally, this can be expressed as:
\begin{equation*}
    \mathcal{D}_{\text{retrieval}} = \{d \in \mathcal{D} : \text{dist}(\phi(o^d_0, l^d), \phi(o^{\text{test}}_0, l^{\text{test}})) \leq \text{dist}_k\},
\end{equation*}

\noindent where $\text{dist}(\cdot, \cdot)$ is some distance function (e.g., cosine distance), and $\text{dist}_k$ is the $k$th smallest distance. With this in mind, we aim to choose $\phi$ such that $\mathcal{D}_{\text{retrieval}}$ has the same generalization relationship to $\tau^{\text{test}}$ as $\mathcal{D}$. To identify what $\phi$ is suitable for this, let us consider each case for how $\tau^{\text{test}}$ can relate to $\mathcal{D}$:

\setlist[enumerate,1]{label=\textbf{Case \arabic*:}, leftmargin=3.75em}
\begin{enumerate}
    \item If $\tau^{\text{test}}$ is \textbf{ID} w.r.t.\ $\mathcal{D}$, there exists some $d^* \in \mathcal{D}$ as previously defined. For $\tau^{\text{test}}$ to also be \textbf{ID} w.r.t.\ $\mathcal{D}_{\text{retrieval}}$, we want $d^* \in \mathcal{D}_{\text{retrieval}}$. Because its initial observation $o_0^*$ is supported by $p_{\text{test}}(o_0)$, which is narrow, $o_0^*$ should be close to $o_0^{\text{test}}$. Therefore, if $\phi$ is not sensitive to small changes in $o_0$, or changes to $l$ that do not change the task specification, it should produce embeddings that are close in distance for $d^*$ and $\tau^{\text{test}}$, such that $d^* \in \mathcal{D}_{\text{retrieval}}$.
    \item If $\tau^{\text{test}}$ captures \textbf{visual} generalization w.r.t.\ $\mathcal{D}$, this is because of the existence of some $d^v \in \mathcal{D}$ as previously defined. For $\tau^{\text{test}}$ to also capture \textbf{visual} generalization w.r.t.\ $\mathcal{D}_{\text{retrieval}}$, we want $d^v \in \mathcal{D}_{\text{retrieval}}$. For this to happen, we want $\phi$ to assign closer distances for tasks that are \textbf{visual} perturbations than \textbf{behavioral} perturbations.
    \item If $\tau^{\text{test}}$ captures \textbf{behavioral} generalization w.r.t.\ $\mathcal{D}$, then all $d \in \mathcal{D}$ are for \textbf{behavioral} perturbations of $\tau^{\text{test}}$. Therefore, regardless of $\phi$, $\mathcal{D}_{\text{retrieval}}$ will also consist entirely of \textbf{behavioral} perturbations, so this case will always result in a proper $\mathcal{D}_{\text{retrieval}}$.
\end{enumerate}

\smallskip \noindent To meet these requirements, we propose using embeddings from VLA policies that have been trained on large-scale robot datasets. We posit that because VLAs inherit VLM backbones, their embeddings possess invariance to changes in visual observations and language that do not affect task specification or behavior. Also, because they have been trained on diverse robot behavior, their embeddings should be sensitive to changes that necessitate different behavior.

We hypothesize that this property transfers to unseen data more readily than policy performance, since a policy need not succeed on a task for its embeddings to be useful for retrieval. VLAs are thus a convenient, increasingly available source of embeddings that requires no additional training.

\begin{figure*}[th]
    \centering
    \includegraphics[width=\linewidth]{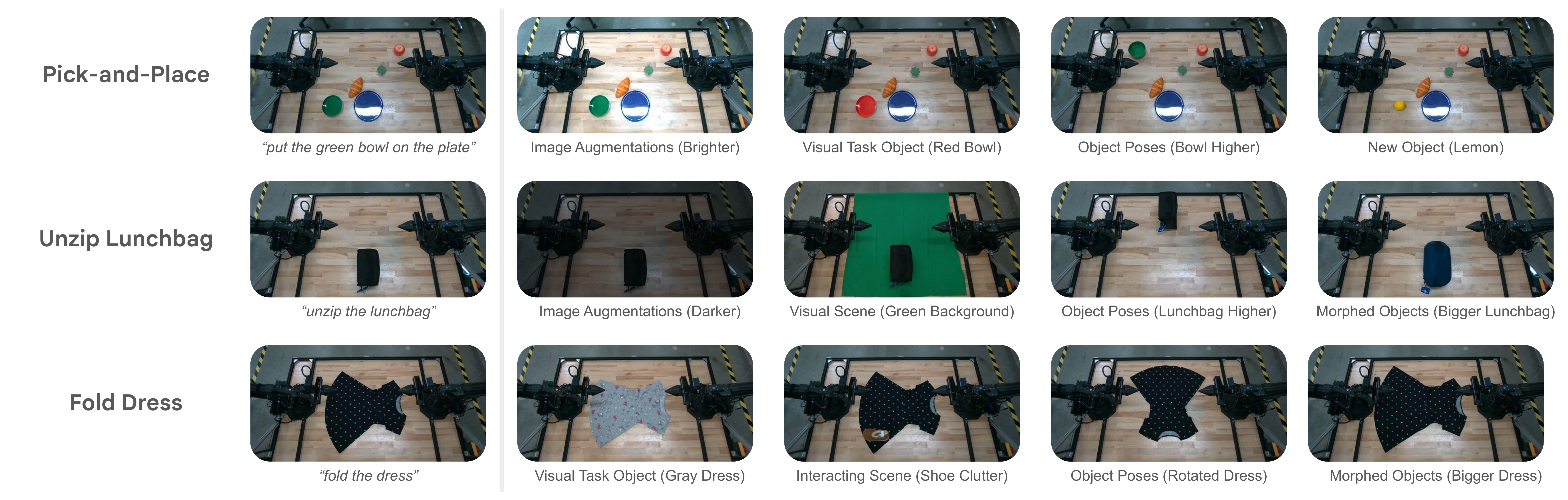}
    \vspace{-10pt}
    \caption{\small We visualize the three task families in our controlled experiments (\emph{Pick-and-Place}, \emph{Unzip Lunchbag}, and \emph{Fold Dress}). We provide an example instance for each task (left), and variations of that task across different axes (right), with the specific change noted in parentheses.}
    \vspace{-5pt}
    \label{fig:task_examples}
\end{figure*}

\subsection{VLM Analysis of Generalization}
\label{sec:vlm_method}
After obtaining $\mathcal{D}_{\text{retrieval}}$, we can use a VLM to analyze it against $\tau^{\text{test}}$. To do this, we prompt it with initial observations $o_0$ and instructions $l$ from each $d \in \mathcal{D}_{\text{retrieval}}$, as well as $o_0^{\text{test}}$ and $l^{\text{test}}$ from $\tau^{\text{test}}$. Then, we ask it to perform pairwise analysis between each $d$ and $\tau^{\text{test}}$, predicting whether $d$ is for a \textbf{visual} perturbation of $\tau^{\text{test}}$ (i.e., $d^v$), a \textbf{behavioral} perturbation, or neither (i.e., $d^*$). Then, we aggregate these predictions to provide an overall classification of $\tau^{\text{test}}$ w.r.t.\ $\mathcal{D}_{\text{retrieval}}$ (and by proxy $\mathcal{D}$), according to the definitions in \cref{sec:setup}.

Before providing a classification for $d$, we ask the VLM to provide qualitative descriptions of how $\tau^{\text{test}}$ differs from $d$ along a variety of different axes from $\bigstar$-Gen. The VLM also provides binary predictions on whether each axis is ID for $\tau^{\text{test}}$ w.r.t.\ $d$. This gives the VLM a prior on what it should reason about when determining how to classify a given example. Also, this makes the VLM output more interpretable, in terms of understanding what task differences resulted in $\tau^{\text{test}}$ being classified a certain way.

\begin{figure*}[t]
    \centering
    \includegraphics[width=0.95\linewidth]{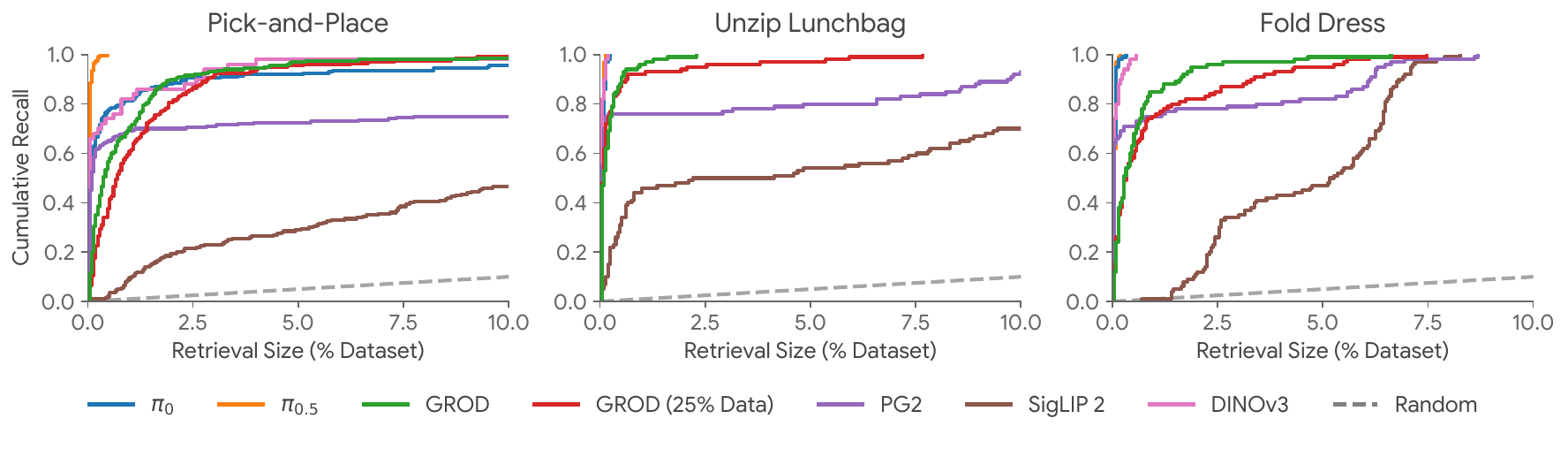}
    \vspace{-8pt}
    \caption{\small Recall rate scaling ($y$-axis) of $d^*$ for $\mathcal{D}_{\text{ID}}$  as the size of the retrieval set ($x$-axis) increases. We find that VLA-based embeddings and DINOv3 significantly outperform the other retrieval methods.}
    \label{fig:in_dist_results}
\end{figure*}

We also consider first asking the VLM to detect all relevant objects in the scene for $\tau^{\text{test}}$ and all $d \in \mathcal{D}_{\text{retrieval}}$, predicting their labels and 2D bounding boxes, before making the other aforementioned predictions. We hypothesize that this can improve predictions on axes that involve object-centric reasoning.

\section{Evaluation}
\label{sec:controlled}
\begin{figure*}[th]
    \centering
    \vspace{-5pt}
    \includegraphics[width=0.95\linewidth]{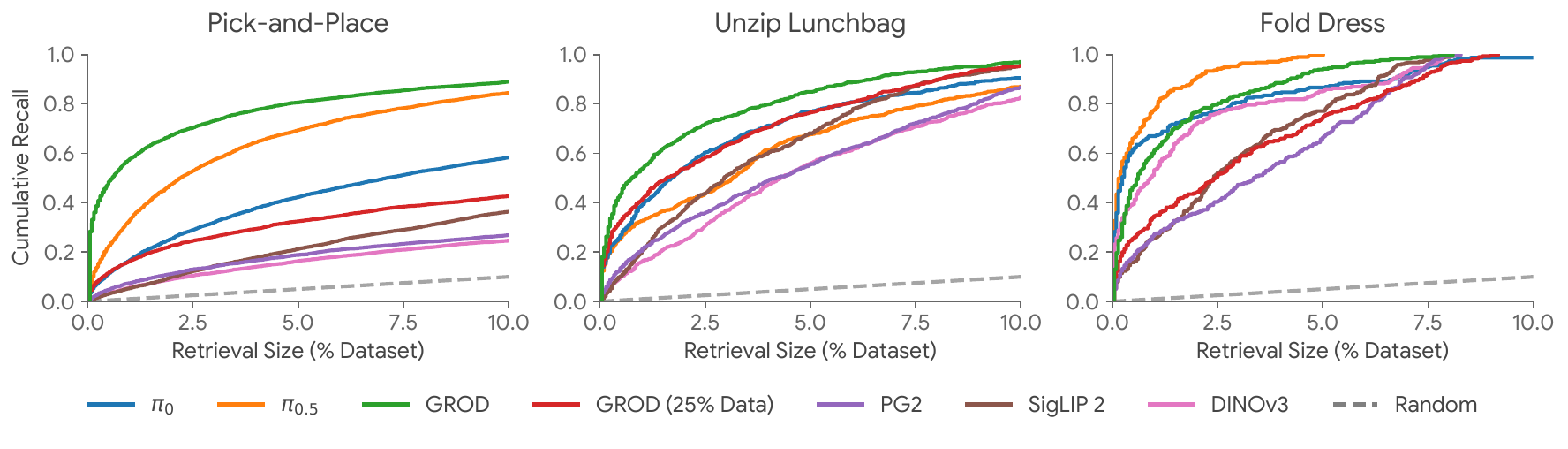}
    \vspace{-8pt}
    \caption{\small Recall rate scaling ($y$-axis) of $d^v$ for $\mathcal{D}_{\text{visual}}$ as the size of the retrieval set ($x$-axis) increases. $\pi_{0.5}$ and GROD outperform their counterparts trained on less robot data ($\pi_{0}$, GROD (25\% Data)), and are much better than the baselines that do not use robot data.}
    \vspace{-10pt}
    \label{fig:visual_results}
\end{figure*}

\begin{figure*}[th]
    \centering
    \begin{subfigure}[b]{\textwidth}
        \centering
        \includegraphics[scale=0.6]{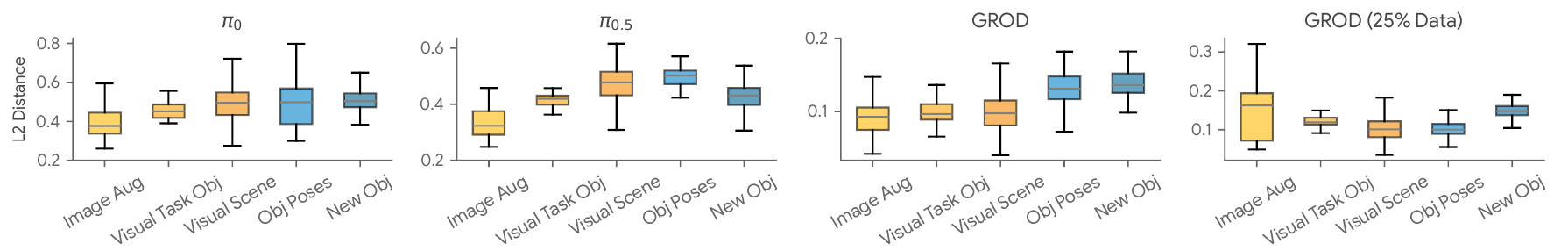}
    \end{subfigure}

    \begin{subfigure}[b]{\textwidth}
        \centering\includegraphics[scale=0.6]{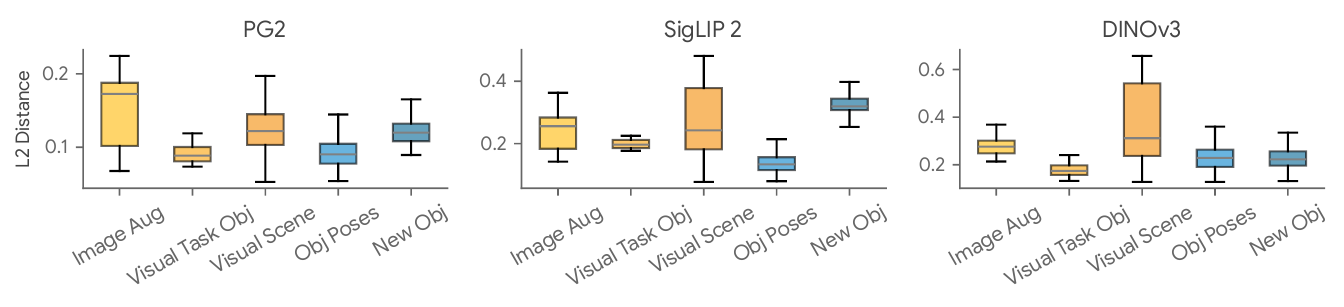}
    \end{subfigure}
    \vspace{-15pt}
    \caption{\small Distribution of embedding distances (L2 with unit-normalized embeddings, equivalent to cosine distance) induced by task perturbations across different axes for \emph{Pick-and-Place}. VLA embeddings, especially for GROD, generally have larger distances for \textbf{behavioral} axes (blue) than \textbf{visual} axes (orange), unlike the other embeddings.}
    \vspace{-10pt}
    \label{fig:axis_dists}
\end{figure*}

To evaluate \method, we conduct controlled experiments where we have access to ground truth to compare against. We instantiate these experiments on the ALOHA 2 platform~\cite{aldaco2024aloha}.

\smallskip \noindent \textbf{Tasks and Datasets.} We collect data for three task families (\emph{Pick-and-Place}, \emph{Unzip Lunchbag}, and \emph{Fold Dress}). Each task family consists of variations for multiple axes, which are annotated to have ground truth labels for evaluation. We visualize some examples of our task variations in \cref{fig:task_examples}. %
In total, our data consists of over 2.3K task variations, which are unseen by the VLAs whose embeddings we use for retrieval.

\smallskip \noindent \textbf{Experimental Setup.}
For each task family, we reserve 50--100 variations as held-out evaluation tasks (i.e., $\tau^{\text{test}}$). Then, for each $\tau^{\text{test}}$, we construct different $\mathcal{D}$, for each way $\tau^{\text{test}}$ can relate to it ($\mathcal{D}_{\text{ID}}$, $\mathcal{D}_{\text{visual}}$, $\mathcal{D}_{\text{behavioral}}$). We consider multiple versions of $\mathcal{D}_{\text{ID}}$ and $\mathcal{D}_{\text{visual}}$ with different $d^*$ and $d^v$, respectively. To prevent trivial retrieval with $\mathcal{D}_{\text{ID}}$, we add minor perturbations to the brightness and language instructions of $d^*$ compared to $\tau^{\text{test}}$ (otherwise, their embeddings would be identical). %
Each $\mathcal{D}$ consists of data across all task families. %

\subsection{Retrieval}
We first compare the effectiveness of retrieval methods. To do this, we plot how the recall rate of relevant examples ($d^*$ for $\mathcal{D}_{\text{ID}}$ and $d^v$ for $\mathcal{D}_{\text{visual}}$) scales with the size of the retrieval set $k$, aggregated across all $\tau^{\text{test}}$ and examples of $d^*$/$d^v$.
Note that we do not consider $\mathcal{D}_{\text{behavioral}}$ because retrieval does not matter for this case. We focus on using \emph{off-the-shelf} embeddings (e.g., no domain-specific training). We compare the following:
\begin{itemize}
    \item $\bm{\pi_0}$/$\bm{\pi_{0.5}}$: We consider the $\pi_0$~\cite{black2024pi_0} and $\pi_{0.5}$~\cite{intelligence2025pi} base VLAs. For embeddings, we use the KV cache corresponding to their image tokens from the second-to-last transformer layer of their PaliGemma~\cite{beyer2024paligemma} backbones.
    \item \textbf{GROD}: We use Gemini Robotics On-Device (GROD)~\cite{DeepMind_GeminiRoboticsOnDevice_2025}, a VLA trained on thousands of hours of ALOHA 2 data. The model is conditioned on four camera views, proprioception, and a language instruction. We extract embeddings similarly as with $\pi_0$/$\pi_{0.5}$. %
    \item \textbf{GROD (25\% Data)}: We consider another version of GROD, but trained on only 25\% of the robot data, scaled down to only data for the most common scenes.
    \item \textbf{PG2}: We use embeddings from PaliGemma 2~\cite{steiner2024paligemma}, where conditioning and embedding extraction are done similarly as with GROD. This comparison ablates the contribution of robot data, compared to only VLM training.
    \item \textbf{SigLIP 2}: We use SigLIP 2~\cite{tschannen2025siglip} embeddings of visual observations and language instructions (concatenated).
    \item \textbf{DINOv3}: We use DINOv3~\cite{simeoni2025dinov3} embeddings of visual observations. We do not incorporate language instructions.
\end{itemize}
We use cosine distance (equivalently, L2 with unit-normalized embeddings) for retrieval. For KV-cache embeddings, we first reduce their dimensionality to 2048 using Gaussian random projections. We also compare against random sampling.

\smallskip \noindent \textbf{In-Distribution Retrieval.} In \cref{fig:in_dist_results}, we report recall rate scaling of $d^*$ for $\mathcal{D}_{\text{ID}}$. We find that all VLA-based embeddings and DINOv3 perform well, achieving recall rates of over 95\% for each task family when retrieving less than 5\% of the original dataset. $\pi_{0.5}$ performs the best overall, particularly on \emph{Pick-and-Place}, which we suspect is because it was trained with knowledge insulation~\cite{driess2025knowledge} to better preserve semantic invariance to the minor language variations in our $d^*$ examples.

Performance degrades slightly with GROD (25\% Data), while the other methods (PG2, SigLIP 2) perform significantly worse. We hypothesize this is because these models lack robustness to the language variations in our $d^*$ examples. While they incorporate language pre-training, they were trained on less language-annotated robot data, or none at all, resulting in worse semantic invariance specific to robot tasks.

\begin{table*}[th]
    \centering
    \small
    \vspace{5pt}
    \resizebox{\textwidth}{!}{%
    \begin{tabular}[t]{l|l|cccccccc|ccc|c}
        \toprule
        Task & VLM & \makecell{Image\\ Aug} & \makecell{Visual\\Task Obj} & \makecell{Visual\\Scene} & \makecell{Obj\\Poses} & \makecell{Morphed\\Obj} & \makecell{New\\Obj} & \makecell{Interacting\\Scene} & \makecell{Perturb\\Class} & \makecell{$\mathcal{D}_{\text{ID}}$\\Acc} & \makecell{$\mathcal{D}_{\text{visual}}$\\Acc} & \makecell{$\mathcal{D}_{\text{behavioral}}$\\Acc} & \makecell{Overall\\Acc} \\
        \midrule
         \multirow{3}{30pt}{\emph{Pick-and-Place}}
         & Gemini 3.1 Flash-Lite & 39.6 & \textbf{81.9} & 39.7 & 29.3 & -- & 87.5 & -- & 78.9 & 95.0 & 7.0 & 78.0 & 60.0 \\
         & Gemini 3 Flash & \textbf{94.7} & 67.3 & 92.7 & 38.2 & -- & 97.5 & -- & 96.7 & \textbf{95.6} & 64.4 & 97.9 & 86.2 \\
         & Gemini 3.1 Pro & 79.1 & 69.4 & \textbf{98.9} & \textbf{73.9} & -- & \textbf{99.4} & -- & \textbf{98.3} & 94.9 & \textbf{82.7} & \textbf{100.0} & \textbf{92.5} \\
       \midrule
         \multirow{3}{30pt}{\emph{Unzip Lunchbag}}
         & Gemini 3.1 Flash-Lite & 52.8 & 84.9 & 75.6 & 75.0 & 55.3 & -- & -- & 83.4 & \textbf{100.0} & 4.0 & \textbf{82.0} & 62.0 \\
         & Gemini 3 Flash & \textbf{95.4} & 96.6 & 89.5 & 81.2 & 65.8 & -- & -- & 88.3 & 95.9 & 58.3 & 51.0 & 68.5 \\
         & Gemini 3.1 Pro & 74.8 & \textbf{97.3} & \textbf{100.0} & \textbf{88.8} & \textbf{69.5} & -- & -- & \textbf{89.7} & 93.8 & \textbf{71.4} & 54.0 & \textbf{72.8} \\
       \midrule
         \multirow{3}{30pt}{\emph{Fold Dress}} 
         & Gemini 3.1 Flash-Lite & 53.4 & 38.4 & 34.1 & 19.8 & \textbf{14.0} & -- & 64.0 & \textbf{81.6} & 92.0 & 8.0 & \textbf{64.0} & 54.7 \\
         & Gemini 3 Flash & \textbf{92.9} & \textbf{99.1} & 67.7 & \textbf{66.6} & 3.6 & -- & \textbf{84.5} & 76.3 & \textbf{100.0} & \textbf{36.2} & 57.5 & \textbf{64.3} \\
         & Gemini 3.1 Pro & 78.2 & 93.7 & \textbf{73.9} & 51.8 & 13.9 & -- & 76.2 & 67.7 & 98.0 & 33.3 & 33.3 & 55.2 \\
        \bottomrule
    \end{tabular}
    }
    \caption{\small (Left) VLM performance (F1 \texttimes100) on pairwise prediction of per-axis differences and perturbation classification between $\tau^{\text{test}}$ and $d \in \mathcal{D}_{\text{retrieval}}$. We do not report axes for which the task has no variation. (Right) Overall \method classification accuracy after aggregating VLM predictions across all $d \in \mathcal{D}_{\text{retrieval}}$, for each version of $\mathcal{D}$.}
    \label{tab:method_results}
\end{table*}

\begin{figure*}[t]
    \centering
    \vspace{-2.5pt}
    \includegraphics[width=\linewidth]{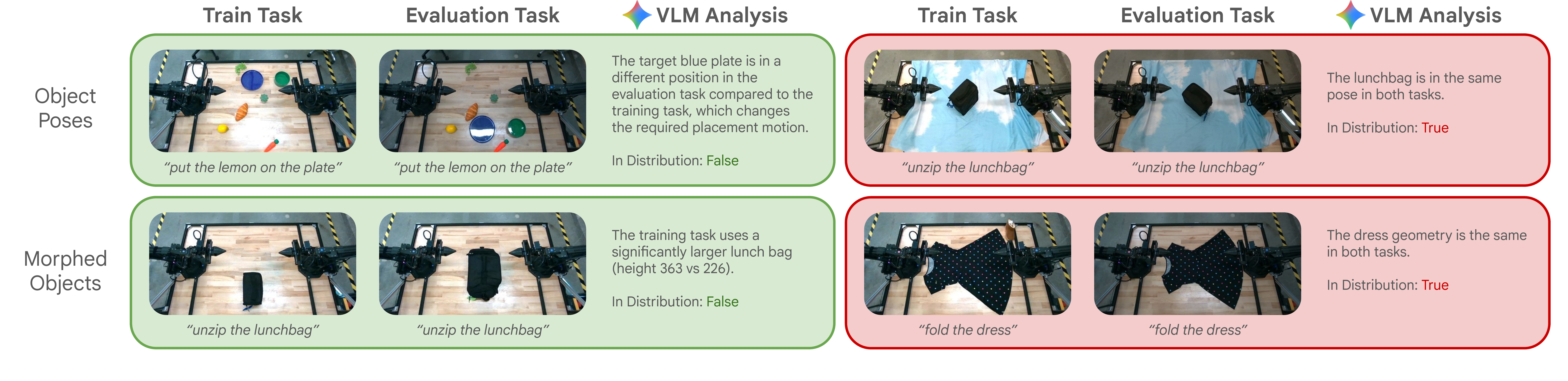}
    \vspace{-17.5pt}
    \caption{\small Examples of successful (green) and failed (red) predictions of Gemini 3 Flash for ``Object Poses'' and ``Morphed Objects''. The VLM struggles more when comparing examples with more subtle visual changes (lunchbag rotation, slightly different dress size).} 
    \vspace{-7.5pt}
    \label{fig:vlm_examples}
\end{figure*}

\smallskip \noindent \textbf{Visual Generalization Retrieval.} In \cref{fig:visual_results}, we report recall rate scaling of $d^v$ for $\mathcal{D}_{\text{visual}}$. We find that GROD performs the best overall, achieving the highest recall rates for 2/3 of our task families. $\pi_{0.5}$ is the only other competitive model, and we suspect GROD performs better because it was trained specifically on data for the evaluation platform (ALOHA 2). However, $\pi_{0.5}$ still performs reasonably, suggesting that more general-purpose VLA embeddings can also be effective.

GROD (25\% Data) and $\pi_0$ perform worse than their counterparts trained on more robot data, indicating that advancements in VLAs, particularly in robot data diversity, can make their embeddings more effective for retrieval. The VLA embeddings significantly outperform the baselines that are not behavior-aware (PG2, SigLIP 2, DINOv3).

While recall is not perfect, we note that this is the recall rate of \emph{individual} examples of $d^v$. If $\mathcal{D}$ contains multiple $d^v$, then retrieval only needs to obtain one. Also, in our experimental setup a large proportion of each dataset is closely related to $\tau^{\text{test}}$, which makes it harder to disambiguate $d^v$.

\smallskip \noindent \textbf{Investigating Embedding Spaces.} To better understand what is captured by each embedding space, we consider how task perturbations for different $\bigstar$-Gen axes affect embedding distances. In \cref{fig:axis_dists}, we visualize box plots of embedding distances induced by perturbations across different axes for \emph{Pick-and-Place}. We show \textbf{visual} axes in shades of orange, and \textbf{behavioral} axes in shades of blue. We see that VLA embeddings, especially for GROD, are generally more sensitive to \textbf{behavioral} perturbations than \textbf{visual}. In contrast, there is more overlap between \textbf{visual} and \textbf{behavioral} axes for the others, explaining why they are less effective for retrieval.

\subsection{VLM Analysis}
Next, we benchmark the VLM analysis and overall performance of \method. We evaluate on retrieval sets $\mathcal{D}_{\text{retrieval}}$ of size $k=20$ obtained using GROD embeddings, which performed the best overall. Each original $\mathcal{D}_{\text{ID}}$ contains two possible $d^*$, while each $\mathcal{D}_{\text{visual}}$ contains up to five possible $d^v$. On these datasets, over 96\% of the resulting retrieval sets contain at least one needed example ($d^*$ or $d^v$, respectively).

We compare %
three different state-of-the-art VLMs in our experiments: Gemini 3.1 Flash-Lite~\cite{gemini31_lite_2026}, 3 Flash~\cite{gemini3_2025}, and 3.1 Pro~\cite{gemini31_2026}. We assess predicting the per-axis differences and perturbation classifications for individual $d \in \mathcal{D}_{\text{retrieval}}$, reporting F1 scores. We also report the overall generalization classification accuracy of \method on $\tau^{\text{test}}$ after aggregating the perturbation classifications across $d$, for each version of $\mathcal{D}$. This accounts for possible error introduced by retrieval.

We report our results in  \cref{tab:method_results}. %
For \emph{Pick-and-Place}, we see positive scaling with stronger VLMs. Gemini 3.1 Pro performs the best, achieving an average F1 score of 0.841 across all per-axis predictions, and an overall classification accuracy of 92.5\%. %
For the other task families, the VLMs struggle more on axes prediction, particularly for ``Object Poses'' and ``Morphed Objects'', resulting in worse perturbation and overall generalization accuracies. We hypothesize this is because axes perturbations for these tasks are more visually subtle than in \emph{Pick-and-Place}. 

\begin{table*}[th]
    \centering
    \small
    \vspace{5pt}
    \resizebox{\textwidth}{!}{%
    \begin{tabular}[t]{l|l|cccccc|ccc|c}
        \toprule
        Task & Method & \makecell{Image\\ Aug} & \makecell{Visual\\Task Obj} & \makecell{Visual\\Scene} & \makecell{Obj\\Poses} & \makecell{Morphed\\Obj} & \makecell{Perturb\\Class} & \makecell{$\mathcal{D}_{\text{ID}}$\\Acc} & \makecell{$\mathcal{D}_{\text{visual}}$\\Acc} & \makecell{$\mathcal{D}_{\text{behavioral}}$\\Acc} &\makecell{Overall\\Acc} \\
        \midrule
         \multirow{4}{25pt}{\emph{Unzip Lunchbag}} & Gemini 3.1 Pro & 74.8 & 97.3 & \textbf{100.0} & \textbf{88.8} & \textbf{69.5} & \textbf{89.7} & 93.8 & \textbf{71.4} & \textbf{54.0} & \textbf{72.8} \\
         & -- Obj Detect & \textbf{75.5} & \textbf{97.7} & 89.9 & 86.7 & 66.0 & 87.2 & 93.9 & 55.1 & 33.3 & 61.0 \\
         & -- Obj Detect, Axes & -- & -- & -- & -- & -- & 79.9 & 90.0 & 40.0 & 6.0 & 45.3 \\
         & + Longer Context ($k=40$) & 73.1 & 95.4 & 99.4 & 87.8 & 68.1 & \textbf{89.7} & \textbf{96.0} & 62.5 & 44.0 & 67.6  \\
        \bottomrule
    \end{tabular}
    }
    \caption{\small Ablations of VLM analysis in \method. (Left) VLM performance (F1 \texttimes100) on pairwise prediction of per-axis differences and perturbation classification between $\tau^{\text{test}}$ and $d \in \mathcal{D}_{\text{retrieval}}$. (Right) Overall \method classification accuracy after aggregating VLM predictions.}
    \label{tab:ablation_results}
    \vspace{-5pt}
\end{table*}

In \cref{fig:vlm_examples}, we visualize examples of successes and failures of 3 Flash, and note that its failure modes typically involve less visually apparent changes. This highlights deficiencies in modern VLMs that can sometimes limit the effectiveness of \method. However, we note that 3.1 Pro does improve for these axes, providing evidence that as VLMs improve these deficiencies can be mitigated. 

\begin{figure*}[th]
    \centering
    \includegraphics[width=\linewidth]{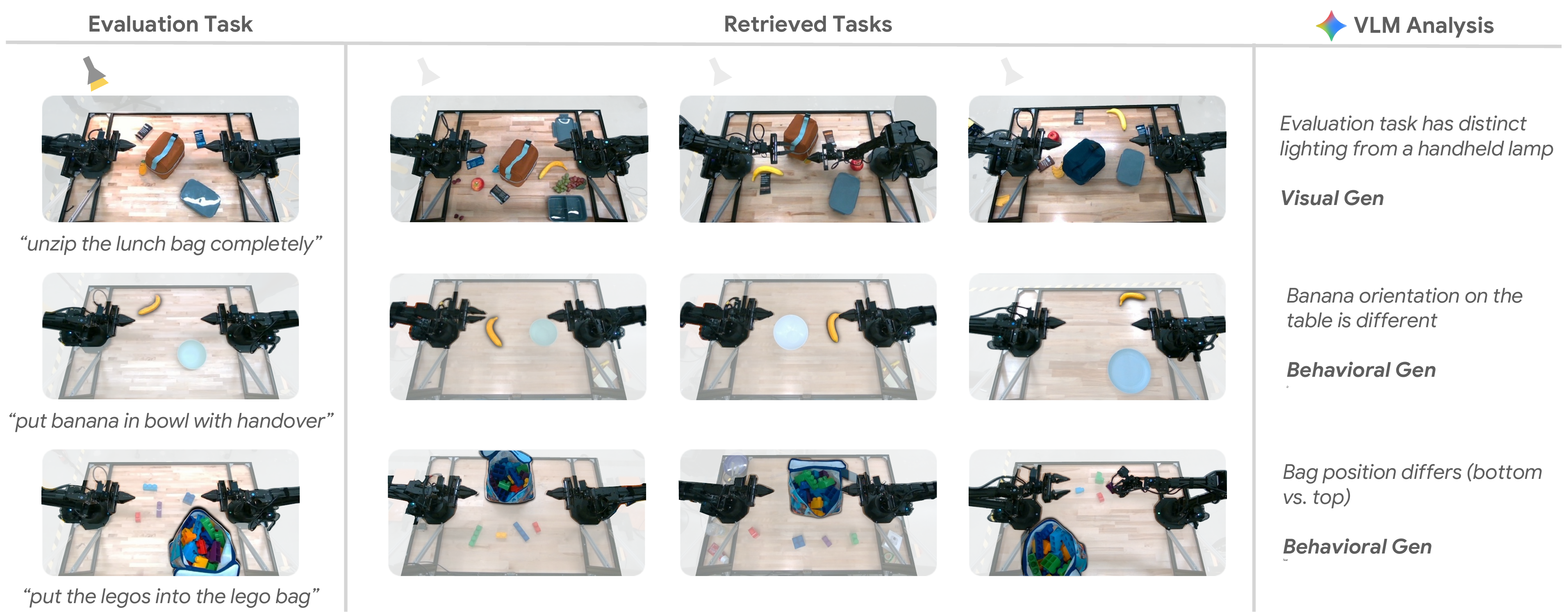}
    \caption{\small Examples of \method applied to benchmarking generalization for a large-scale ALOHA 2 dataset. For a given benchmark evaluation task (left), we provide three instances of retrieved training data examples (middle). Then, we provide the VLM analysis comparing the evaluation task to the leftmost retrieved example, as well as the overall generalization classification (right).}
    \vspace{-10pt}
    \label{fig:bench_examples}
\end{figure*}

\smallskip \noindent \textbf{Ablations.} We ablate design decisions in our instantiation of VLM analysis in \method. Using Gemini 3.1 Pro, we compare our full approach on \emph{Unzip Lunchbag} to omitting object detections, as well as also omitting $\bigstar$-Gen axes guidance. In addition, we increase the size of the retrieval set from $k=20$ to $40$, to assess how sensitive \method is to longer contexts. %

In our results in \cref{tab:ablation_results}, we find that including object detection slightly improves object-centric axes (``Object Poses'', ``Morphed Objects''), as well as ``Visual Scene''. This translates to an overall accuracy improvement of around 12\%.
When removing $\bigstar$-Gen axes guidance, generalization classification accuracy significantly drops. This shows the importance of using $\bigstar$-Gen to guide VLM reasoning, so that it can effectively determine what forms of generalization are involved.

Lastly, when increasing the size of the retrieval set, we find that axes predictions degrade slightly, suggesting that VLMs perform worse at these forms of visual reasoning with longer contexts. This results in somewhat worse overall classification accuracies, although performance does not degrade substantially, indicating some robustness to context length.

\section{Case Study: Large-Scale Analysis}
\label{sec:large_scale}
We now show how \method can be applied to large-scale robot datasets. We consider Bridge V2~\cite{walke2023bridgedata}, and a proprietary dataset of over 1M demonstrations for the ALOHA 2 platform.

\smallskip \noindent \textbf{Bridge V2.} We consider 35 evaluation tasks from BridgeV2-$\bigstar$~\cite{taxonomy2025arxiv}, a benchmark from prior work used to assess generalization with Bridge V2. We use \method to analyze these tasks against the entirety of Bridge V2, as well as the base tasks used for co-training policies in the BridgeV2-$\bigstar$ evaluations. We use retrieval size $k=50$ with embeddings from CogACT~\cite{li2024cogact}, as this VLA is more suited for the Bridge V2 platform compared to the VLAs in our previous experiments.

In \cref{tab:bridge_results}, we report the overall generalization classification agreement of different VLMs in \method, compared against the original categorizations in BridgeV2-$\bigstar$, and see agreement improves with stronger VLMs. We observe that disagreements are largely due to visually subtle changes, similar to the VLM errors in our controlled experiments.

While retrieved Bridge V2 data is generally relevant (similar scenes and objects), none captured the exact behavior of a given evaluation task; all were thus categorized as \textbf{behavioral}, leaving only the co-training base tasks relevant for categorization. This results in significant alignment between \method and the categorizations from BridgeV2-$\bigstar$, which originally defined generalization with respect to these base tasks.

\begin{table}[t]
    \centering
    \small
    \vspace{5pt}
    \resizebox{\columnwidth}{!}{%
    \begin{tabular}[t]{lccc}
        \toprule
        VLM & Gemini 3 Flash & Gemini 3.0 Pro & Gemini 3.1 Pro  \\
        \midrule
        Agreement & 60.0 & 71.4 & 77.1 \\
        \bottomrule
    \end{tabular}
    }
    \caption{\small Classification agreement of different VLMs in \method on BridgeV2-$\bigstar$~\cite{taxonomy2025arxiv}, compared against their original categorizations.}
    \label{tab:bridge_results}
    \vspace{-15pt}
\end{table}

\smallskip \noindent \textbf{ALOHA 2.} While the ALOHA 2 dataset does not have ground truth labels to evaluate against, we can assess \method by considering evaluation tasks from prior work that were used to benchmark generalization for policies trained on this data~\cite{team2025gemini}. %
We can then compare \method analysis against how these evaluations were originally characterized. We use retrieval size $k=50$ with GROD embeddings, and Gemini 3.1 Pro.

In \cref{fig:bench_examples}, we provide examples where \method agrees with previous human characterizations of the aforementioned benchmark tasks, including retrieved examples, VLM analysis, and overall generalization classification. However, we also observe that sometimes there is disagreement, e.g., sometimes humans appear in the background of scenes that were categorized as \textbf{in-distribution}, but \method identifies this instead as \textbf{visual} generalization. This shows how \method can be used to systematically identify limitations in previous benchmarking efforts that may otherwise have been overlooked.

\section{Discussion}
\label{sec:discussion}
We present \method, a scalable framework for characterizing the extent and type of generalization required for policy evaluation. \method involves a two-stage pipeline where a retrieval step using VLA embeddings first identifies relevant training data. Then, a VLM analyzes the retrieved data against the evaluation task, providing comparisons across various generalization axes and an overall generalization classification.

In our experiments, we show that VLA embeddings perform well at identifying examples that enable accurate downstream analysis, and VLMs can be effective at predicting axis differences and classifying generalization. We also scale \method to two large-scale robot datasets, where it produces analysis consistent with human-defined generalization benchmarks. %

\smallskip \noindent \textbf{Downstream Applications.} Because \method characterizes the generalization required by an evaluation relative to training data, its outputs can inform several practical decisions: auditing whether a benchmark measures the generalization it intends, revealing coverage gaps to guide data collection and model improvement, and informing deployment decisions about which conditions a policy is likely to handle. Prior work such as $\bigstar$-Gen demonstrates the value of analyzing policy performance against generalization type, but obtains this characterization through manual benchmark design, whereas \method produces it automatically and at scale.

\smallskip \noindent \textbf{Limitations and Future Work.} While we find that VLMs are effective at analyzing \emph{Pick-and-Place}, they perform worse on our other tasks, particularly with more subtle visual changes. We expect that this will improve as VLMs continue to advance.

\method assumes that task differences are observable from their initial observations and instructions, which may not always be true. In particular, behavioral differences that only emerge during execution (e.g., in high-precision insertion) can be difficult to detect from an initial observation, and distinguishing them from visually similar in-distribution examples grows harder as datasets scale and neighbors become closer. Furthermore, task differences may be more challenging to disambiguate if observations across tasks are very misaligned (e.g., different camera views). Future work can address this limitation by incorporating 3D scene understanding, which may require further advancements in VLMs.

While \method can help characterize generalization, its level of categorization is relatively broad (\textbf{in-distribution}, \textbf{visual}, or \textbf{behavioral} generalization). Future work can investigate more granular analysis, such as estimating the difficulty of generalization required (e.g., whether a change requires significantly different behavior, or only a minor adjustment). %

\bibliographystyle{IEEEtran}
{
\footnotesize
\setlength{\bibsep}{0pt plus 0.3ex}
\bibliography{references}
}

\clearpage
\appendices
\label{sec:appendix}
\section{Experimental Details}

\subsection{Hyperparameter Summary}
\label[appendix]{sec:hyperparams}
For reproducibility, we consolidate the key hyperparameters and inputs for both stages of \method in \cref{tab:hyperparams}. The full prompt templates, including the per-neighbor input format and the complete list of $\bigstar$-Gen axes with their definitions, are provided in \cref{fig:axes_prompt_part1,fig:obj_prompt_part2,fig:no_obj_prompt_part2,fig:no_axes_prompt}, and the aggregation procedure is described in \cref{alg:vlm_analysis}.

\begin{table}[t]
\centering
\footnotesize
\setlength{\tabcolsep}{4pt}
\renewcommand{\arraystretch}{1.2}
\begin{tabular}{@{}p{0.32\columnwidth} p{0.6\columnwidth}@{}}
\toprule
\multicolumn{2}{@{}l}{\textbf{Retrieval}} \\
\midrule
Embedding feature & KV cache of image tokens from the 2nd-to-last transformer backbone layer (PaliGemma VLAs); cognition feature (CogACT) \\
Raw dimensionality & 131072 ($\pi_0$, $\pi_{0.5}$); 4096 (CogACT) \\
Dim.\ reduction & Gaussian random projection to 2048 (KV-cache embeddings) \\
Distance metric & Cosine \\
Embedding inputs & VLA embeddings: camera views, language instruction $l$, and 14-d proprioception (per arm: 6-d joint position + 1-d gripper position, as in ALOHA); CogACT omits proprioception. SigLIP~2/DINOv3 embed the overhead view only (DINOv3 omits language) \\
Retrieval size $k$ & 20 (main); 40 (long-context ablation); 50 (large-scale case studies) \\
\midrule
\multicolumn{2}{@{}l}{\textbf{VLM Analysis}} \\
\midrule
Temperature & 0 \\
Thinking budget & High \\
Output format & Enforced JSON schema \\
Per-neighbor inputs & Task ID, language instruction, single overhead RGB scene image \\
Eval-task inputs & Language instruction, single overhead RGB scene image \\
Visual axes & Image Augmentations, Visual Task Object, Visual Scene \\
Behavioral axes & Object Poses, Morphed Objects, New Object, Interacting Scene \\
Additional axis & ``Other'' (differences not covered by the above) \\
Object detection (optional) & Per-object label + 2D bounding box [$y_{\min}, x_{\min}, y_{\max}, x_{\max}$] normalized to 0--1000, for the eval task and each neighbor \\
Aggregation & Per-axis ID/OOD $\rightarrow$ per-neighbor label $\rightarrow$ overall classification (\cref{alg:vlm_analysis}) \\
\bottomrule
\end{tabular}
\caption{Summary of retrieval and VLM analysis hyperparameters and inputs for \method.}
\label{tab:hyperparams}
\end{table}

\subsection{Embeddings}
For $\pi_0$ and $\pi_{0.5}$, we obtain embeddings using the KV cache corresponding to image tokens from the second-to-last transformer layer of their PaliGemma backbones. This results in an embedding with dimensionality 131072.

For CogACT, we use the 4096-dimensional ``cognition feature'' produced by its VLM backbone to condition its action expert.

For SigLIP 2, we use the Hugging Face model \texttt{google/siglip2-giant-opt-patch16-384}.

For DINOv3, we use \\
\texttt{facebook/dinov3-vitl16-pretrain-lvd1689m}. For SigLIP 2 and DINOv3, we only embed the overhead camera view, rather than all camera views as for the VLA embeddings.

\subsection{VLM Analysis Details}
\label[appendix]{sec:vlm_details}
In \cref{fig:axes_prompt_part1}, we provide Part 1 of the VLM prompt we use when incorporating $\bigstar$-Gen axes guidance. In \cref{fig:obj_prompt_part2}, we provide Part 2 of this prompt when also using object detection, and in \cref{fig:no_obj_prompt_part2} we provide Part 2 when not using object detection. In \cref{fig:no_axes_prompt}, we provide the full VLM prompt we use when not using $\bigstar$-Gen axes guidance or object detection.

For Gemini, we set temperature to 0 and set thinking to \texttt{high}. Also, we enforce the response to follow a JSON schema for more reliable generation and extraction of per-axis predictions and generalization classifications.

When analyzing $\tau^{\text{test}}$ against a particular $d$, we ask the VLM to explicitly predict a perturbation classification. However, if also predicting $\bigstar$-Gen axes differences, we instead use these axes predictions to construct the perturbation classification rather than the explicitly predicted one, as we found this generally performed better, especially for Gemini 3 Flash.

\cref{alg:vlm_analysis} describes the procedure for producing a generalization classification using $\bigstar$-Gen guidance. In Step 1, the VLM does a pairwise comparison between the evaluation task and each retrieved task to determine whether they are in-distribution for each $\bigstar$-Gen axis. These axes predictions are used to assign a perturbation classification for the retrieved task. Then, in Step 2, the classifications for all retrieved tasks are aggregated into an overall generalization classification.

\begin{algorithm}
\caption{VLM Generalization Categorization}
\begin{algorithmic}[1]
\Require Eval task $\tau^{\text{test}}$; retrieval set $\mathcal{D}_{\text{retrieval}}=\{d_1,\dots,d_k\}$;
     axis sets $A_{\text{behavioral}}$, $A_{\text{visual}}$
\Statex \textbf{Step 1: categorize each retrieved task}
\For{each retrieved task $d_i \in \mathcal{D}_{\text{retrieval}}$}
  \State $z_i \gets \textsc{PairwiseCompare}(\tau^{\text{test}}, d_i)$ \Comment{$z_{i,a}=1$ iff in-distribution on axis $a$}
  \If{$\exists\, a \in A_{\text{behavioral}} : z_{i,a}=0$}
      \State $\ell_i \gets \texttt{behavioral}$
  \ElsIf{$\exists\, a \in A_{\text{visual}} : z_{i,a}=0$}
      \State $\ell_i \gets \texttt{visual}$
  \Else
      \State $\ell_i \gets \texttt{in\text{-}distribution}$
  \EndIf
\EndFor
\Statex \textbf{Step 2: aggregate into overall prediction}
\State $\hat{y} \gets \texttt{behavioral}$
\If{$\exists\, i : \ell_i = \texttt{in\text{-}distribution}$}
  \State $\hat{y} \gets \texttt{in\text{-}distribution}$
\ElsIf{$\exists\, i : \ell_i = \texttt{visual}$}
  \State $\hat{y} \gets \texttt{visual}$
\EndIf
\State \Return $\hat{y}$
\end{algorithmic}
\label{alg:vlm_analysis}
\end{algorithm}

\begin{figure*}
\begin{tcolorbox}[colback=gray!5, colframe=gray!75]
A robot is given a task, which consists of a physical scene it interacts with, and a language instruction that specifies what behavior the robot needs to perform. To perform the task, the robot must predict actions to execute given image observations of its scene and the language instruction. The robot may encounter tasks that represent generalization from the tasks in its training data. Tasks can require visual generalization, which means the robot does not need to generalize to new physical motion, but must be robust to visual perturbations. An example of this is performing a task present in the training data, but with unseen lighting conditions that affect the robot's visual observations. Tasks can also require behavioral generalization, which means the robot must execute new physical motion that is unseen in its training data. An example of this is grasping an object with unseen geometry that requires a new grasp motion. \\

To help categorize what kind of generalization a task represents, consider the following axes that tasks can differ, grouped based on whether they involve visual or behavioral generalization: \\
**Visual** \\
- Image Augmentations: Changes to the scene image observation that does not affect the physical composition of the scene (e.g., scene lighting, image blur). Pay attention to any such visual changes in the image observations. \\
- Visual Task Object: Changes to the appearance (e.g., color, visual texture, transparency) of objects that are involved in the task (e.g., an object to be grasped, a container an object is to be put in). This does not include aspects of objects that also affect their physical configuration (e.g., pose, geometry). \\
- Visual Scene: Changes to visual elements of the scene besides objects involved in the task (e.g., the color of a tabletop surface, distractor objects that do not affect the required task behavior). \\
**Behavioral** \\
- Object Poses: Changes to the pose (position and orientation) of objects involved in the task that affect the required task behavior. \\
- Morphed Objects: Changes to the geometry (size and shape) of objects involved in the task that affect the required task behavior. This does not include changes to the type of an object (e.g., this includes changing the shape of a cup, but not changing the cup to a bottle). \\
- New Object: Changes to objects involved in the task to entirely new types of objects with different appearances and physical characteristics, such as geometry, that affect the required task behavior. \\
- Interacting Scene: Changes to other physical components of the scene besides objects involved in the task that affect the required task behavior (e.g., tabletop surface height, object clutter that blocks objects involved in the task). This only includes changes that affect the required task behavior (e.g., object clutter is only included if the change forces the robot to perform different behavior, such as to avoid the objects). \\

The robot is trained on data for the following tasks, each represented by its language instruction and initial scene image observation: \\

Task ID: 1 \\
Instruction: \{retrieved instruction 1\} \\
Scene Image: \{retrieved image 1\} \\

Task ID: 2 \\
Instruction: \{retrieved instruction 2\} \\
Scene Image: \{retrieved image 2\} \\

... \\

After being trained on data for these tasks, the robot is asked to perform the following evaluation task: \\
Instruction: \{test instruction\} \\
Scene Image: \{test image\} \\

\end{tcolorbox}
\caption{VLM prompt with $\bigstar$-Gen axes guidance (Part 1).}
\label{fig:axes_prompt_part1}
\end{figure*}

\begin{figure*}
\begin{tcolorbox}[colback=gray!5, colframe=gray!75]
Your goal is to analyze how the evaluation task represents generalization from the training tasks. First, detect all objects in the evaluation task scene, providing a label and bounding box (formatted as [y\_min, x\_min, y\_max, x\_max], with coordinates normalized between 0-1000) for each object. Also, for each of the training task, detect all objects in the same format. Next, for each training task, provide its Task ID, and then a qualitative description of how the evaluation task and the training task differ for each axis, and whether or not the evaluation task is in-distribution with respect to the training task for that axis (True or False). If there is a difference between the tasks for that axis, then it is not in-distribution. Also do this for an ``Other'' axis that captures any other ways the tasks can differ that are not covered by the other axes. Use the object detections to help reason about these differences. When analyzing behavioral axes, pay especially close attention to any differences, even minor ones, that can affect the actions needed to perform the task. Consider estimating the surface area of objects involved in the task to help with analyzing behavioral axes. Also, pay attention to the language instruction for each task when making comparisons, as the instruction can change what objects are involved in each task (e.g., the same object can appear in different tasks, but may only be involved in one depending on the instruction). Then, use this to categorize the evaluation task as either in-distribution, visual generalization, or behavioral generalization for each chosen training task. If some behavioral axes are not in-distribution, then the task is behavioral generalization, even if some visual axes are also not in-distribution. If the only axes that are not in-distribution are visual, then the task is visual generalization. If all axes are in-distribution, then the task is in-distribution. IMPORTANT: Make sure to provide an analysis for every training task, do not skip any. Provide your response in JSON format.
\end{tcolorbox}
\caption{VLM prompt with $\bigstar$-Gen axes guidance and object detection (Part 2).}
\label{fig:obj_prompt_part2}
\end{figure*}

\begin{figure*}
\begin{tcolorbox}[colback=gray!5, colframe=gray!75]
Your goal is to analyze how the evaluation task represents generalization from the training tasks. For each of the training tasks, provide a qualitative description of how the evaluation task and the training task differ for each axis, and whether or not the evaluation task is in-distribution with respect to the training task for that axis (True or False). If there is a difference between the tasks for that axis, then it is not in-distribution. Also do this for an ``Other'' axis that captures any other ways the tasks can differ that are not covered by the other axes. When analyzing behavioral axes, pay especially close attention to any differences, even minor ones, that can affect the actions needed to perform the task. Consider estimating the surface area of objects involved in the task to help with analyzing behavioral axes. Also, pay attention to the language instruction for each task when making comparisons, as the instruction can change what objects are involved in each task (e.g., the same object can appear in different tasks, but may only be involved in one depending on the instruction). Then, use this to categorize the evaluation task as either in-distribution, visual generalization, or behavioral generalization for each chosen training task. If some behavioral axes are not in-distribution, then the task is behavioral generalization, even if some visual axes are also not in-distribution. If the only axes that are not in-distribution are visual, then the task is visual generalization. If all axes are in-distribution, then the task is in-distribution. IMPORTANT: Make sure to provide an analysis for every training task, do not skip any. Provide your response in JSON format.
\end{tcolorbox}
\caption{VLM prompt with $\bigstar$-Gen axes guidance, but without object detection (Part 2).}
\label{fig:no_obj_prompt_part2}
\end{figure*}

\begin{figure*}
\begin{tcolorbox}[colback=gray!5, colframe=gray!75]
A robot is given a task, which consists of a physical scene it interacts with, and a language instruction that specifies what behavior the robot needs to perform. To perform the task, the robot must predict actions to execute given image observations of its scene and the language instruction. The robot may encounter tasks that represent generalization from the tasks in its training data. Tasks can require visual generalization, which means the robot does not need to generalize to new physical motion, but must be robust to visual perturbations. An example of this is performing a task present in the training data, but with unseen lighting conditions that affect the robot's visual observations. Tasks can also require behavioral generalization, which means the robot must execute new physical motion that is unseen in its training data. An example of this is grasping an object with unseen geometry that requires a new grasp motion. \\

The robot is trained on data for the following tasks, each represented by its language instruction and initial scene image observation: \\

Task ID: 1 \\
Instruction: \{retrieved instruction 1\} \\
Scene Image: \{retrieved image 1\} \\

Task ID: 2 \\
Instruction: \{retrieved instruction 2\} \\
Scene Image: \{retrieved image 2\} \\

... \\

After being trained on data for these tasks, the robot is asked to perform the following evaluation task: \\
Instruction: \{test instruction\} \\
Scene Image: \{test image\} \\

Your goal is to analyze how the evaluation task represents generalization from the training tasks. Categorize the evaluation task as either in-distribution, visual generalization, or behavioral generalization for each training task. IMPORTANT: Make sure to provide an analysis for every training task, do not skip any. Provide your response in JSON format.

\end{tcolorbox}
\caption{VLM prompt without $\bigstar$-Gen axes guidance or object detection.}
\label{fig:no_axes_prompt}
\end{figure*}

\subsection{Tasks and Datasets}
\label[appendix]{sec:task_details}
\noindent \textbf{Pick-and-Place.} This task setup involves a plate in the center of a table. There is one object to the left of the plate, and one object to the right. The robot is instructed to place one of these two objects on the plate. There are 4 different possible manipulated objects (green bowl, red bowl, red apple, yellow lemon). There are also two different poses on the plate's left and two different poses on the plate's right for these objects. In addition, there are 3 different configurations of distractor objects (including none), and 2 different plate poses. We collect scenes that cover every possible combination of this variability, which results in a total of $\binom{4}{2}$ (which objects are chosen) $\times$ 2 (which object is on the left, right) $\times$ 2 (left object pose) $\times$ 2 (right object pose) $\times$ 3 (distractor configuration) $\times$ 2 (plate pose) = 288 unique scenes. Each scene has 2 language instructions (one for the left object, one for the right object), for a total of 576 unique tasks.

We consider task instances to be OOD with respect to each other for different axes as follows:
\begin{itemize}
    \item \textbf{Visual Task Object}: The color of the manipulated object differs.
    \item \textbf{Visual Scene}: The configuration of non-manipulated objects in the scenes differs.
    \item \textbf{Object Poses}: The pose of the manipulated object or the plate differs.
    \item \textbf{New Object}: The type of manipulated object differs (bowl vs. apple vs. lemon).
\end{itemize}

\smallskip \noindent \textbf{Unzip Lunchbag.} This task setup involves 6 different possible lunchbags. 3 of them have the same geometry, but have visual differences (tag or bag color). The other 3 have significantly different geometries. We further vary the task by having 4 different table backgrounds and 5 different lunchbag poses. This results in a total of 6 $\times$ 4 $\times$ 5 = 120 unique scenes/tasks.

We consider task instances to be OOD with respect to each other for different axes as follows:
\begin{itemize}
    \item \textbf{Visual Task Object}: The color of the lunchbag or tag differs.
    \item \textbf{Visual Scene}: The table background differs.
    \item \textbf{Object Poses}: The pose of the lunchbag differs.
    \item \textbf{Morphed Objects}: The geometry of the lunchbag differs.
\end{itemize}

\smallskip \noindent \textbf{Fold Dress.} This task setup involves 4 different dresses. 3 of them have the same visual design but are of different sizes, while the other has the same geometry as another but has a different visual design. We further vary the task by having 4 different dress orientations, and 5 different configurations of other objects (this includes no other objects, 2 configurations with distractor objects that do not affect task behavior, and 2 with clutter where objects are placed on the dress). This results in a total of 4 $\times$ 4 $\times$ 5 = 80 unique scenes/tasks.

We consider task instances to be OOD with respect to each other for different axes as follows:
\begin{itemize}
    \item \textbf{Visual Task Object}: The color of the dress differs.
    \item \textbf{Visual Scene}: The distractor objects differ.
    \item \textbf{Object Poses}: The pose of the dress differs.
    \item \textbf{Morphed Objects}: The geometry of the dress differs.
    \item \textbf{Interacting Scene}: The clutter objects differ.
\end{itemize}

After collecting data for each scene/task, we create 2 additional variants of each with different levels of image brightness. This corresponds to differences in the axis ``Image Augmentations''. This results in a total of (576 + 120 + 80) $\times$ 3 = 2328 task variations.

\subsection{Experimental Setup}
\label[appendix]{sec:setup_details}
For each task family, we randomly reserve 50 scenes as held-out for evaluation. This results in 100 evaluation tasks for \emph{Pick-and-Place}, because there are two possible tasks per scene, and 50 for the other task families. Then separately for each evaluation task $\tau^{\text{test}}$, we construct different versions of $\mathcal{D}_{\text{ID}}$, $\mathcal{D}_{\text{visual}}$,  and $\mathcal{D}_{\text{behavioral}}$ as follows:

\begin{itemize}
    \item For $\mathcal{D}_{\text{ID}}$, we start with every non-evaluation task across all task family datasets. Then, for the given $\tau^{\text{test}}$, we include two different versions of $d^*$, which have the same image observation, but with small random changes to their brightness. Each also has their language instruction slightly modified (e.g., ``put the red apple on the plate'' $\rightarrow$ ``put the red apple on the dish'', ``unzip the lunchbag'' $\rightarrow$ ``open the lunchbag''). These modifications are introduced such that in-distribution retrieval is not trivial, as if $d^*$ has identical observations and instructions as $\tau^{\text{test}}$, then any function would work for retrieval (same input $\rightarrow$ same output).
    \item For $\mathcal{D}_{\text{visual}}$, we start with every non-evaluation task that involves different behavior from $\tau^{\text{test}}$ across all task family datasets. We then sample and include up to 5 examples of $d^v$ that involve the same behavior as $\tau^{\text{test}}$. We sample fewer than 5 if fewer than 5 exist.
    \item For $\mathcal{D}_{\text{behavioral}}$, we use every non-evaluation task that involves different behavior from $\tau^{\text{test}}$ across all task family datasets.
\end{itemize}

For our retrieval recall scaling results (\cref{fig:in_dist_results} and \cref{fig:visual_results}), we do not explicitly construct these datasets. Instead, we start with the versions of each dataset without $d^*$/$d^v$, and sort them according to their embedding distances. Then, for every possible $d^*$ and $d^v$ (not just 5 samples of $d^v$, but all examples in the dataset that could serve as $d^v$), we determine where in the sorted order they would fall, which is the retrieval size that would be needed to recover it. Then, in the recall scaling plots, each point on a curve indicates what percent of $d^*$/$d^v$ are found in a retrieval set of that size relative to $\mathcal{D}_{\text{ID}}$/$\mathcal{D}_{\text{visual}}$, across all $d^*$/$d^v$ for all evaluation tasks.

\section{Additional Results}

\subsection{Set-Level Retrieval Success}
\label[appendix]{sec:set_level}
Our main retrieval results (\cref{fig:in_dist_results,fig:visual_results}) report the recall rate of \emph{individual} relevant examples ($d^*$ or $d^v$). In practice, however, \method only requires that the retrieval set $\mathcal{D}_{\text{retrieval}}$ contain \emph{at least one} relevant example for VLM analysis. We therefore also report a set-level retrieval success rate: the fraction of evaluation tasks for which the top-$k$ retrieval set contains at least one relevant example, as a function of the retrieval set size $k$. We compute this for our two strongest embeddings, GROD and $\pi_{0.5}$, using the same retrieval rankings as \cref{fig:in_dist_results,fig:visual_results} (i.e., considering all candidate $d^*$/$d^v$).

\begin{figure*}[t]
    \centering
    \includegraphics[width=\linewidth]{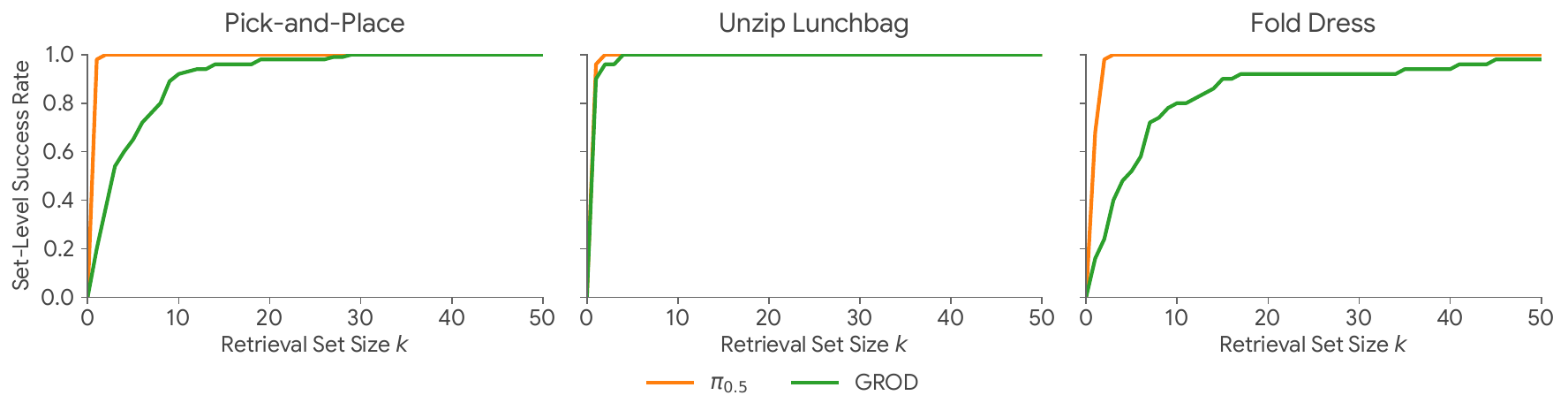}
    \caption{\small Set-level retrieval success for $\mathcal{D}_{\text{ID}}$: the fraction of evaluation tasks whose top-$k$ retrieval set contains at least one $d^*$, as a function of the retrieval set size $k$, for GROD and $\pi_{0.5}$.}
    \label{fig:set_level_in_dist}
\end{figure*}

\begin{figure*}[t]
    \centering
    \includegraphics[width=\linewidth]{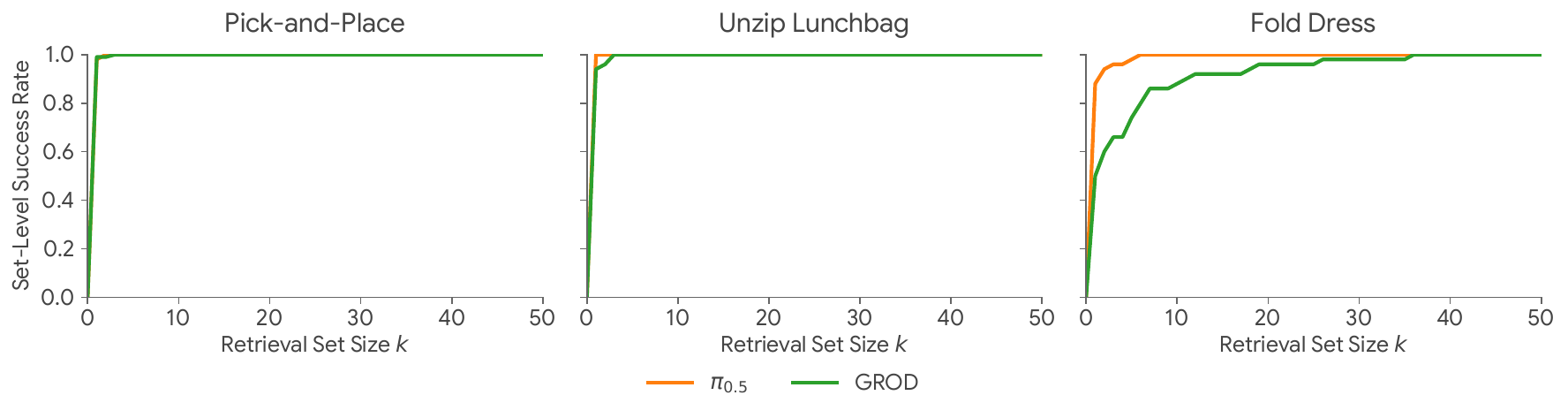}
    \caption{\small Set-level retrieval success for $\mathcal{D}_{\text{visual}}$: the fraction of evaluation tasks whose top-$k$ retrieval set contains at least one $d^v$, as a function of the retrieval set size $k$, for GROD and $\pi_{0.5}$.}
    \label{fig:set_level_visual}
\end{figure*}

We report set-level success for $\mathcal{D}_{\text{ID}}$ in \cref{fig:set_level_in_dist} and for $\mathcal{D}_{\text{visual}}$ in \cref{fig:set_level_visual}. For both embeddings, the retrieval set contains a relevant example with high probability at small $k$: at our default $k=20$, success is at least 97\% for both $\mathcal{D}_{\text{ID}}$ and $\mathcal{D}_{\text{visual}}$. Because only a single relevant example is needed, set-level success saturates much faster than per-example recall, which is why we treat per-example recall as the more informative metric for comparing embeddings in the main text.

\subsection{Qualitative Probes of Retrieval}
\label[appendix]{sec:qual_probes}

We provide two qualitative probes of how our retrieval step behaves under controlled changes via Gemini image editing. Both use a single-view Bridge-like task with CogACT policy embeddings~\cite{li2024cogact}: we use this setup because controlled image edits are only straightforward for single-view scenes, whereas our multi-view ALOHA 2 setting makes consistent edits difficult beyond simple changes such as lighting.

\begin{figure*}[t]
  \centering
  \begin{subfigure}{0.24\textwidth}
    \centering
    \includegraphics[width=\linewidth]{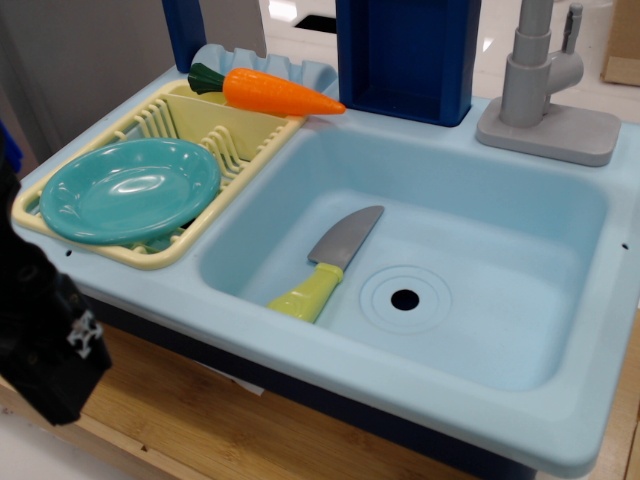}
    \caption{Base}
  \end{subfigure}
  \hfill
  \begin{subfigure}{0.24\textwidth}
    \centering
    \includegraphics[width=\linewidth]{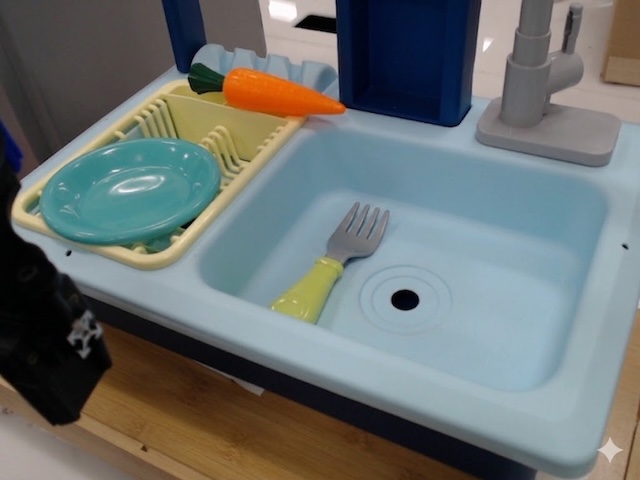}
    \caption{Fork ($d_{\cos}{=}0.021$)}
  \end{subfigure}
  \hfill
  \begin{subfigure}{0.24\textwidth}
    \centering
    \includegraphics[width=\linewidth]{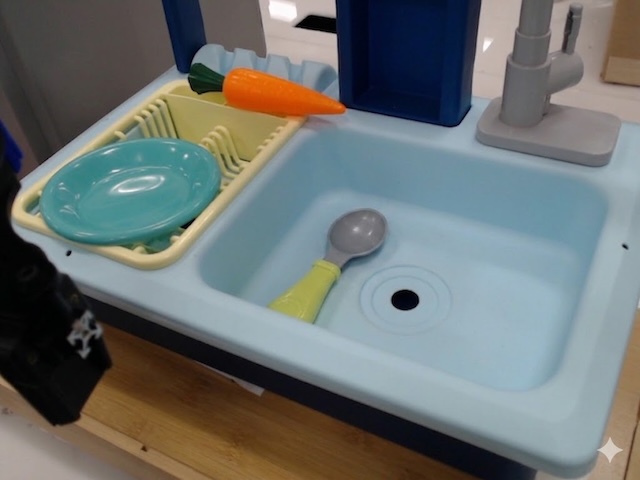}
    \caption{Spoon ($d_{\cos}{=}0.064$)}
  \end{subfigure}
  \hfill
  \begin{subfigure}{0.24\textwidth}
    \centering
    \includegraphics[width=\linewidth]{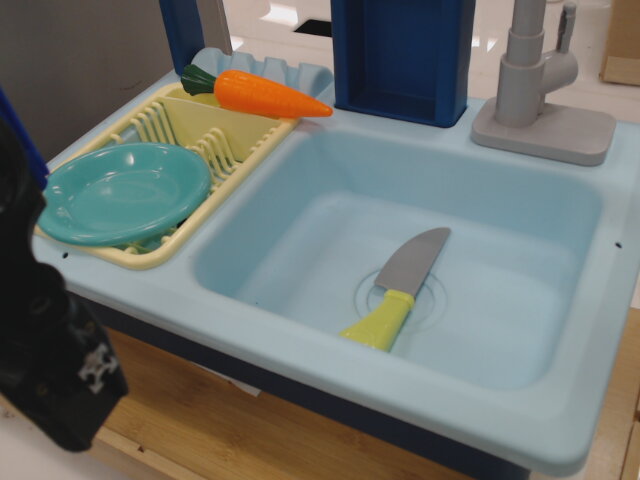}
    \caption{Knife moved ($d_{\cos}{=}0.145$)}
  \end{subfigure}
  \caption{\small Distinct object classes with functionally identical behavior, for the instruction \emph{``put utensil on plate''}. Starting from a base scene with a knife (a), we swap in a fork (b) and spoon (c) while keeping the handle geometry and pose fixed, yielding \textbf{visual} perturbations that require the same behavior. Moving the original knife to a new pose (d) is a \textbf{behavioral} perturbation. We report the cosine distance ($d_{\cos}$) of each image to the base scene in the CogACT policy embedding space used for retrieval. The different-class fork and spoon are closer to the base than the same-class repositioned knife, so embedding proximity tracks the required behavior rather than object-class identity.}
  \label{fig:modified_bridge}
\end{figure*}

\smallskip \noindent \textbf{Distinct object classes with functionally identical behavior.} We first test whether functionally equivalent objects from distinct classes are identified as \textbf{visual} perturbations, using the instruction \emph{``put utensil on plate''} (\cref{fig:modified_bridge}). Starting from a base scene containing a knife, we replace the knife with a fork and a spoon while keeping the handle geometry and pose fixed, so that the optimal behavior for grasping and placing them remains functionally identical. These are thus \textbf{visual} perturbations despite belonging to distinct object classes. For comparison, we construct a \textbf{behavioral} perturbation by moving the original knife to a different pose. Measuring the embedding cosine distance from each modified image to the base scene, the fork and spoon are substantially closer to the base (0.021 and 0.064) than the repositioned knife (0.145). This shows that embedding proximity tracks required behavior rather than only object-class identity.

\begin{figure*}[t]
  \centering
  \captionsetup[sub]{justification=centering}
  \begin{subfigure}[t]{0.135\textwidth}
    \centering
    \includegraphics[width=\linewidth]{figures/modified_bridge/base.jpg}
    \caption{Base}
  \end{subfigure}
  \hfill
  \begin{subfigure}[t]{0.135\textwidth}
    \centering
    \includegraphics[width=\linewidth]{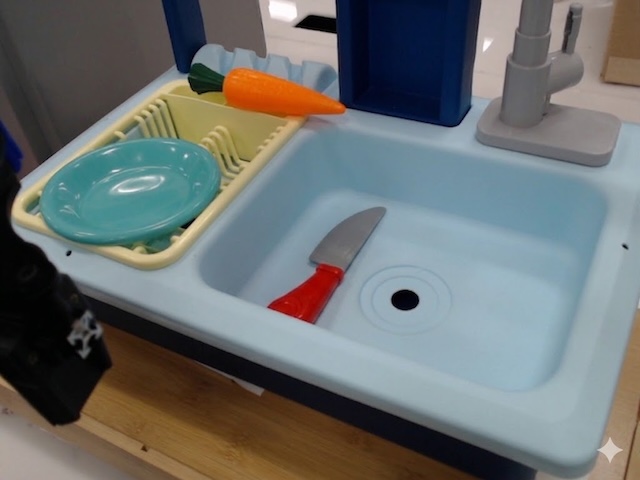}
    \caption{+ Red handle \\ ($d_{\cos}{=}0.037$)}
  \end{subfigure}
  \hfill
  \begin{subfigure}[t]{0.135\textwidth}
    \centering
    \includegraphics[width=\linewidth]{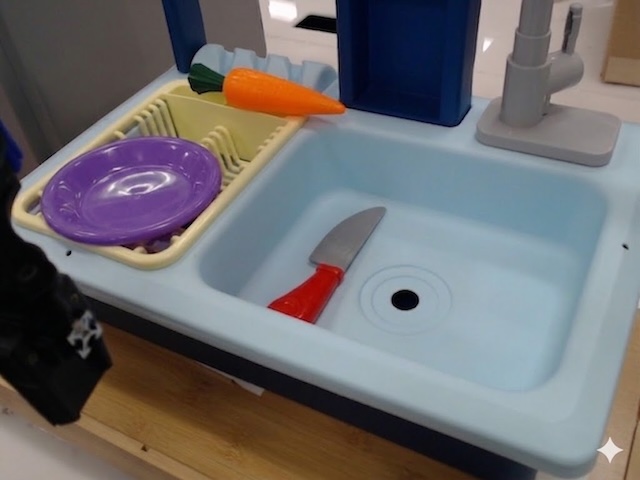}
    \caption{+ Purple plate \\ ($d_{\cos}{=}0.049$)}
  \end{subfigure}
  \hfill
  \begin{subfigure}[t]{0.135\textwidth}
    \centering
    \includegraphics[width=\linewidth]{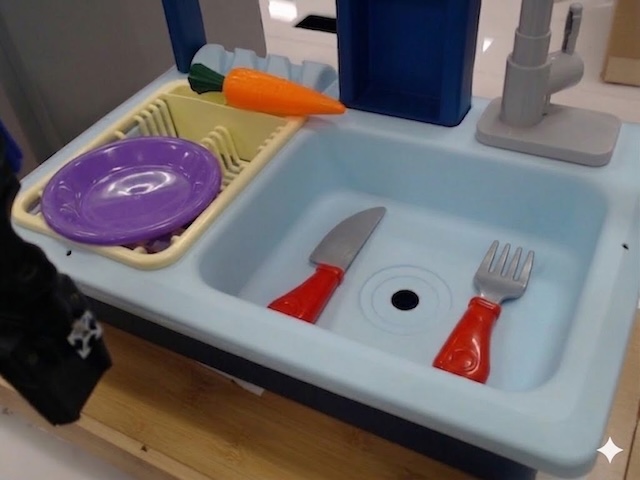}
    \caption{+ Distractor fork \\ ($d_{\cos}{=}0.102$)}
  \end{subfigure}
  \hfill
  \begin{subfigure}[t]{0.135\textwidth}
    \centering
    \includegraphics[width=\linewidth]{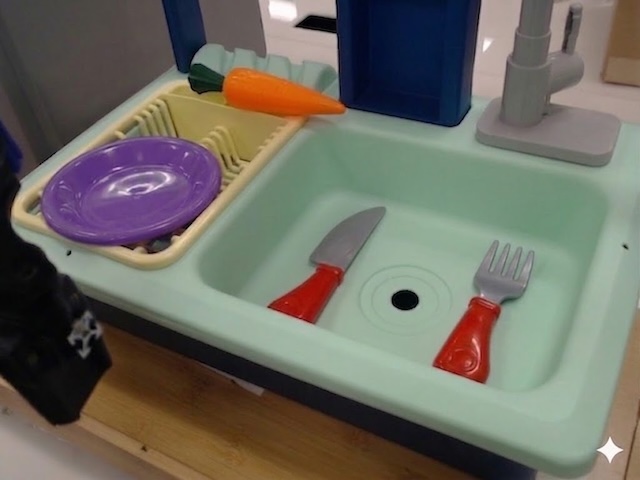}
    \caption{+ Green sink \\ ($d_{\cos}{=}0.122$)}
  \end{subfigure}
  \hfill
  \begin{subfigure}[t]{0.135\textwidth}
    \centering
    \includegraphics[width=\linewidth]{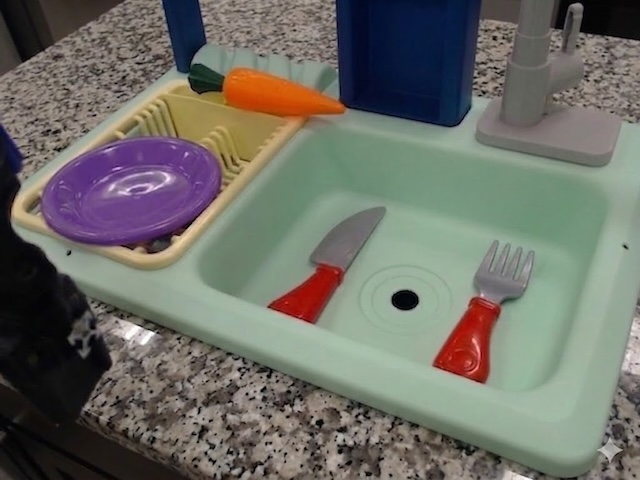}
    \caption{+ Granite counter \\ ($d_{\cos}{=}0.201$)}
  \end{subfigure}
  \hfill
  \begin{subfigure}[t]{0.135\textwidth}
    \centering
    \includegraphics[width=\linewidth]{figures/modified_bridge/knife_right.jpg}
    \caption{Knife moved \\ ($d_{\cos}{=}0.130$)}
  \end{subfigure}
  \caption{\small Stress-testing retrieval under increasing \textbf{visual} shifts, for the instruction \emph{``put knife on plate''}. Starting from the base scene (a), we apply progressively larger, compounding \textbf{visual} changes (b--f), and report their cosine distances to the base in CogACT embedding space. Only the most extreme \textbf{visual} shift (+ granite counter, f) exceeds the embedding distance for the \textbf{behavioral} example (knife moved, g).}
  \label{fig:shift_magnitude}
\end{figure*}

\smallskip \noindent \textbf{Stress-testing retrieval under large visual shifts.} One possible failure mode of our retrieval step is a large \textbf{visual} shift that becomes too distant to recover, causing misclassification as \textbf{behavioral}. We assess this possibility in \cref{fig:shift_magnitude} for the instruction \emph{``put knife on plate''}, where we apply progressively larger, compounding image edits and report their embedding cosine distances to the base scene. The embedding distance for these \textbf{visual} perturbations remains lower than that of a ground-truth \textbf{behavioral} example (knife moved, $d_{\cos}{=}0.130$), except for the most extreme shift (+ granite counter, $d_{\cos}{=}0.201$).

\subsection{Confusion Matrices and Error Analysis}
In \cref{fig:confusion-per-task}, we provide per-task family confusion matrices for our results in \cref{tab:method_results} with Gemini 3.1 Pro. We observe that \textbf{ID} examples are identified reliably everywhere (94--98\% recall), while the two other categories are detected less consistently: 46 \textbf{visual} and 18 \textbf{behavioral} scenarios are misclassified as \textbf{ID}. We hypothesize this bias comes from our aggregation rule, under which a scenario is labeled \textbf{ID} if any retrieved example is predicted as such. \method also confuses the two generalization types asymmetrically, mislabeling 37 \textbf{behavioral} scenarios as \textbf{visual} versus only 17 in the reverse direction. These errors are concentrated in the deformable object tasks: for example, \emph{Fold Dress} has \textbf{visual} and \textbf{behavioral} recall both at chance level (33\%). This suggests that Gemini can reason well about rigid manipulation, but struggles more when judging if task variation with deformable objects demands new physical behavior.

\begin{figure*}[t]
\centering
\begin{subtable}{0.32\textwidth}
\centering
\resizebox{\linewidth}{!}{%
\begin{tabular}{@{}l ccc@{}}
\toprule
& \multicolumn{3}{c}{Predicted} \\
\cmidrule(lr){2-4}
True & $\mathcal{D}_{\text{ID}}$ & $\mathcal{D}_{\text{visual}}$ & $\mathcal{D}_{\text{behavioral}}$ \\
\midrule
$\mathcal{D}_{\text{ID}}$         & \textbf{93} & 4  & 1  \\
$\mathcal{D}_{\text{visual}}$     & 7  & \textbf{81} & 10 \\
$\mathcal{D}_{\text{behavioral}}$ & 0  & 0  & \textbf{99} \\
\bottomrule
\end{tabular}}
\caption{\emph{Pick-and-Place}}
\end{subtable}
\hfill
\begin{subtable}{0.32\textwidth}
\centering
\resizebox{\linewidth}{!}{%
\begin{tabular}{@{}l ccc@{}}
\toprule
& \multicolumn{3}{c}{Predicted} \\
\cmidrule(lr){2-4}
True & $\mathcal{D}_{\text{ID}}$ & $\mathcal{D}_{\text{visual}}$ & $\mathcal{D}_{\text{behavioral}}$ \\
\midrule
$\mathcal{D}_{\text{ID}}$         & \textbf{45} & 1  & 2  \\
$\mathcal{D}_{\text{visual}}$     & 9  & \textbf{35} & 5  \\
$\mathcal{D}_{\text{behavioral}}$ & 2  & 21 & \textbf{27} \\
\bottomrule
\end{tabular}}
\caption{\emph{Unzip Lunchbag}}
\end{subtable}
\hfill
\begin{subtable}{0.32\textwidth}
\centering
\resizebox{\linewidth}{!}{%
\begin{tabular}{@{}l ccc@{}}
\toprule
& \multicolumn{3}{c}{Predicted} \\
\cmidrule(lr){2-4}
True & $\mathcal{D}_{\text{ID}}$ & $\mathcal{D}_{\text{visual}}$ & $\mathcal{D}_{\text{behavioral}}$ \\
\midrule
$\mathcal{D}_{\text{ID}}$         & \textbf{48} & 1  & 0  \\
$\mathcal{D}_{\text{visual}}$     & 30 & \textbf{16} & 2  \\
$\mathcal{D}_{\text{behavioral}}$ & 16 & 16 & \textbf{16} \\
\bottomrule
\end{tabular}}
\caption{\emph{Fold Dress}}
\end{subtable}
\caption{Per-task family RADAR classification confusion matrices using Gemini 3.1 Pro. Rows are ground truth, columns are the predicted overall generalization type; diagonal entries (correct) are \textbf{bold}.}
\label{fig:confusion-per-task}
\end{figure*}

\subsection{Robust Aggregation}
\label[appendix]{sec:robust_agg}
While the aggregation procedure described in \cref{alg:vlm_analysis} is principled assuming that the classifications per retrieved example are correct, it can lack robustness, as a single misclassified example can result in an erroneous overall prediction. Therefore, here we investigate alternate methods of aggregation across the retrieval set, which include the following:
\begin{itemize}
    \item \textbf{Multi-Support}: Rather than requiring only a single \textbf{ID} or \textbf{visual} neighbor to determine a classification, we instead require $m > 1$ (e.g., there must exist $m$ neighbors classified as \textbf{ID} for the overall prediction to be \textbf{ID}). We experiment with $m=2$.
    \item \textbf{Rank-Weighted Voting}: We assign each neighbor a weight proportional to $\frac{1}{R}$, where $R$ is its rank in the retrieval set according to embedding distance ($R=1$ corresponds to the nearest neighbor). We normalize the weights to sum to 1. Then, for a threshold $\theta$, if the sum of weights for examples classified as $\textbf{ID}$ is at least $\theta$, we assign an overall prediction of $\textbf{ID}$. Similarly, if the sum of weights for examples classified as either $\textbf{ID}$ or $\textbf{visual}$ is at least $\theta$, we predict $\textbf{visual}$. Otherwise, we predict $\textbf{behavioral}$. We experiment with $\theta=0.1$.
    \item \textbf{Uncertainty}: In situations where the overall classification decision rests on a single neighbor (e.g., there exists a single neighbor classified as \textbf{ID}), we instead assign a prediction of ``uncertain''.
\end{itemize}

  \begin{table}[th]
      \centering
      \small
      \vspace{5pt}
      \resizebox{\columnwidth}{!}{%
      \begin{tabular}[t]{l|l|ccc|c}
          \toprule
          Task & Aggregation Method & \makecell{$\mathcal{D}_{\text{ID}}$\\Acc} & \makecell{$\mathcal{D}_{\text{visual}}$\\Acc} & \makecell{$\mathcal{D}_{\text{behavioral}}$\\Acc} &
   \makecell{Overall\\Acc} \\
          \midrule
           \multirow{4}{40pt}{\emph{Pick-and-Place}}
           & Original        & \textbf{94.9} & \textbf{82.7} & \textbf{100.0} & \textbf{92.5} \\
           & Multi-Support      & 92.9 & 61.2 & \textbf{100.0} & 84.7 \\
           & Rank-Weighted & 93.9 & 53.1 & \textbf{100.0} & 82.4 \\
           & \emph{Uncertainty (87.8\% cov.)} & \emph{94.8} & \emph{84.4} & \emph{100.0} & \emph{94.2} \\
         \midrule
           \multirow{4}{40pt}{\emph{Unzip Lunchbag}}
           & Original        & \textbf{93.8} & \textbf{71.4} & 54.0 & 72.8 \\
           & Multi-Support      & \textbf{93.8} & 65.3 & 58.0 & 72.1 \\
           & Rank-Weighted & \textbf{93.8} & 67.3 & \textbf{74.0} & \textbf{78.2} \\
           & \emph{Uncertainty (85.7\% cov.)} & \emph{93.8} & \emph{80.6} & \emph{57.4} & \emph{77.0} \\
         \midrule
           \multirow{4}{40pt}{\emph{Fold Dress}}
           & Original        & \textbf{98.0} & 33.3 & 33.3 & 55.2 \\
           & Multi-Support      & 93.9 & \textbf{79.2} & 39.6 & \textbf{71.0} \\
           & Rank-Weighted & 95.9 & 54.2 & \textbf{45.8} & 65.5 \\
           & \emph{Uncertainty (72.4\% cov.)} & \emph{97.9} & \emph{62.5} & \emph{47.1} & \emph{73.3} \\
          \bottomrule
      \end{tabular}
      }
      \caption{\small Comparing methods for aggregating per-neighbor predictions in \method. We compare the \emph{Original} method described in \cref{alg:vlm_analysis} with three alternate methods. \emph{Uncertainty} accuracies (italic) are computed over a subset of examples (coverage shown in parentheses), and are therefore not directly comparable.}
      \label{tab:aggregation_results}
  \end{table}

We summarize the results in \cref{tab:aggregation_results}. The robust aggregation methods slightly hurt \emph{Pick-and-Place}. This is because $\mathcal{D}_{\text{behavioral}}$ accuracy is already perfect, so there is no benefit in requiring redundancy to label an example as not \textbf{behavioral}. However, on the other tasks, the robust aggregation methods improve overall performance. For example, with \emph{Fold Dress}, Multi-Support significantly improves overall accuracy, mostly by recovering $\mathcal{D}_{\text{visual}}$ accuracy. On \emph{Unzip Lunchbag}, Rank-Weighted helps the most, by improving $\mathcal{D}_{\text{behavioral}}$ accuracy. $\mathcal{D}_{\text{ID}}$ accuracy remains high (92--98\%) across all methods, showing that the robust variants trade a small amount of $\mathcal{D}_{\text{ID}}$ sensitivity for large gains on the other categories. Uncertainty achieves the highest accuracy on the scenarios it predicts by abstaining on the 12--28\% of cases whose label rests on a single neighbor, illustrating that much of the residual error is concentrated in inherently ambiguous scenarios.

Overall, no single rule dominates: the original aggregation method is best when per-neighbor VLM classifications are reliable, while the robust aggregation methods are better for the tasks where these classifications are less reliable.

\end{document}